\RequirePackage{snapshot}
\documentclass[letterpaper, 10 pt, conference]{ieeeconf}  
\IEEEoverridecommandlockouts                              
\usepackage{epsfig}
\usepackage{booktabs, multicol, multirow}
\usepackage[export]{adjustbox}
\usepackage{amsfonts,amsmath,amssymb,mathtools,bm}
\usepackage{booktabs}
\usepackage[table]{xcolor}    
\usepackage[font=footnotesize]{caption}
\newcolumntype{P}[1]{>{\centering\arraybackslash}p{#1}}
\newcolumntype{M}[1]{>{\centering\arraybackslash}m{#1}}

\usepackage{pifont}
\newcommand{\cmark}{\text{\ding{51}}}
\newcommand{\xmark}{\text{\ding{55}}}

\usepackage{algorithm}
\usepackage{algpseudocode}
\makeatletter
\let\OldStatex\Statex
\renewcommand{\Statex}[1][0]{%
  \setlength\@tempdima{\algorithmicindent}%
  \OldStatex\hskip\dimexpr#1\@tempdima\relax}
\makeatother

\renewcommand{\algorithmicrequire}{\textbf{Input:}}
\renewcommand{\algorithmicensure}{\textbf{Output:}}
\algnewcommand\AND{~\textbf{and}~}
\algnewcommand\OR{~\textbf{or}~}
\algnewcommand\CONTINUE{~\textbf{continue}~}

\makeatletter
\algnewcommand{\LineComment}[1]{\Statex \hskip\ALG@thistlm \(\triangleright\)
  #1}
\makeatother

\usepackage{times}

\usepackage{flushend}

\definecolor{fullred}{rgb}{0.85,.0,.1} 
\definecolor{navyblue}{rgb}{.0,.0,.5}
\definecolor{bleudefrance}{rgb}{0.19, 0.55, 0.91}
\definecolor{bluegray}{rgb}{0.18, 0.36, 0.6}
\definecolor{lightgray}{rgb}{0.4, 0.4, 0.4}

\makeatletter
\let\NAT@parse\undefined
\makeatother

\usepackage{texscripts}

\newcommand{\vx}{\bm{x}}

\newcommand{\Bx}{\mathbf{x}}

\newcommand{\Bz}{\mathbf{z}}

\usepackage[pagebackref=true,breaklinks=true,letterpaper=true,colorlinks,bookmarks=false,urlcolor=navyblue]{hyperref}

\begin{document}

\title{\LARGE \bf Towards Visual Ego-motion Learning in Robots}

\author{Sudeep Pillai\\
CSAIL, MIT\\
{\tt\small \href{mailto:spillai@csail.mit.edu}{spillai@csail.mit.edu}}
\and
John J. Leonard\\
MIT\\
{\tt\small \href{mailto:jleonard@mit.edu}{jleonard@mit.edu}}
\thanks{Sudeep Pillai and John J. Leonard are with the Computer
Science and Artificial Intelligence Lab (CSAIL), Massachusetts
Institute of Technology (MIT), Cambridge MA 02139, USA. This work was
partially supported by the Office of Naval Research under grants
N00014-11-1-0688 and N00014-13-1-0588 and by the National Science
Foundation under grant IIS-1318392, which we gratefully
acknowledge. For more details,
visit~\url{http://people.csail.mit.edu/spillai/learning-egomotion}.}
}

\maketitle
\thispagestyle{empty}

\begin{abstract} Many model-based Visual Odometry (VO) algorithms have
been proposed in the past decade, often restricted to the type of
camera optics, or the underlying motion manifold observed. We envision
robots to be able to learn and perform these tasks, in a minimally
supervised setting, as they gain more experience. To this end, we
propose a fully trainable solution to visual ego-motion estimation for
varied camera optics. We propose a visual ego-motion learning
architecture that maps observed optical flow vectors to an ego-motion
density estimate via a Mixture Density Network (MDN). By modeling the
architecture as a Conditional Variational Autoencoder (C-VAE), our
model is able to provide introspective reasoning and prediction for
ego-motion induced scene-flow. Additionally, our proposed model is
especially amenable to \textit{bootstrapped} ego-motion learning in
robots where the supervision in ego-motion estimation for a particular
camera sensor can be obtained from standard navigation-based sensor
fusion strategies (GPS/INS and wheel-odometry fusion). Through
experiments, we show the utility of our proposed approach in enabling
the concept of self-supervised learning for visual ego-motion
estimation in autonomous robots.



\end{abstract}

\section{Introduction}\label{sec:introduction} Visual odometry
(VO)~\cite{nister2004visual}, commonly referred to as ego-motion
estimation, is a fundamental capability that enables robots to reliably
navigate its immediate environment. With the wide-spread adoption of
cameras in various robotics applications, there has been an evolution
in visual odometry algorithms with a wide set of variants including
monocular VO~\cite{nister2004visual,konolige2010large}, stereo
VO~\cite{howard2008real,kitt2010visual} and even non-overlapping
\textit{n}-camera VO~\cite{hee2013motion,kneip2013using}. Furthermore, each of
these algorithms has been custom tailored for specific camera optics
(pinhole, fisheye, catadioptric) and the range of motions observed by
these cameras mounted on various
platforms~\cite{scaramuzza20111}.

\begin{figure}[t]
  \includegraphics[width=\columnwidth]{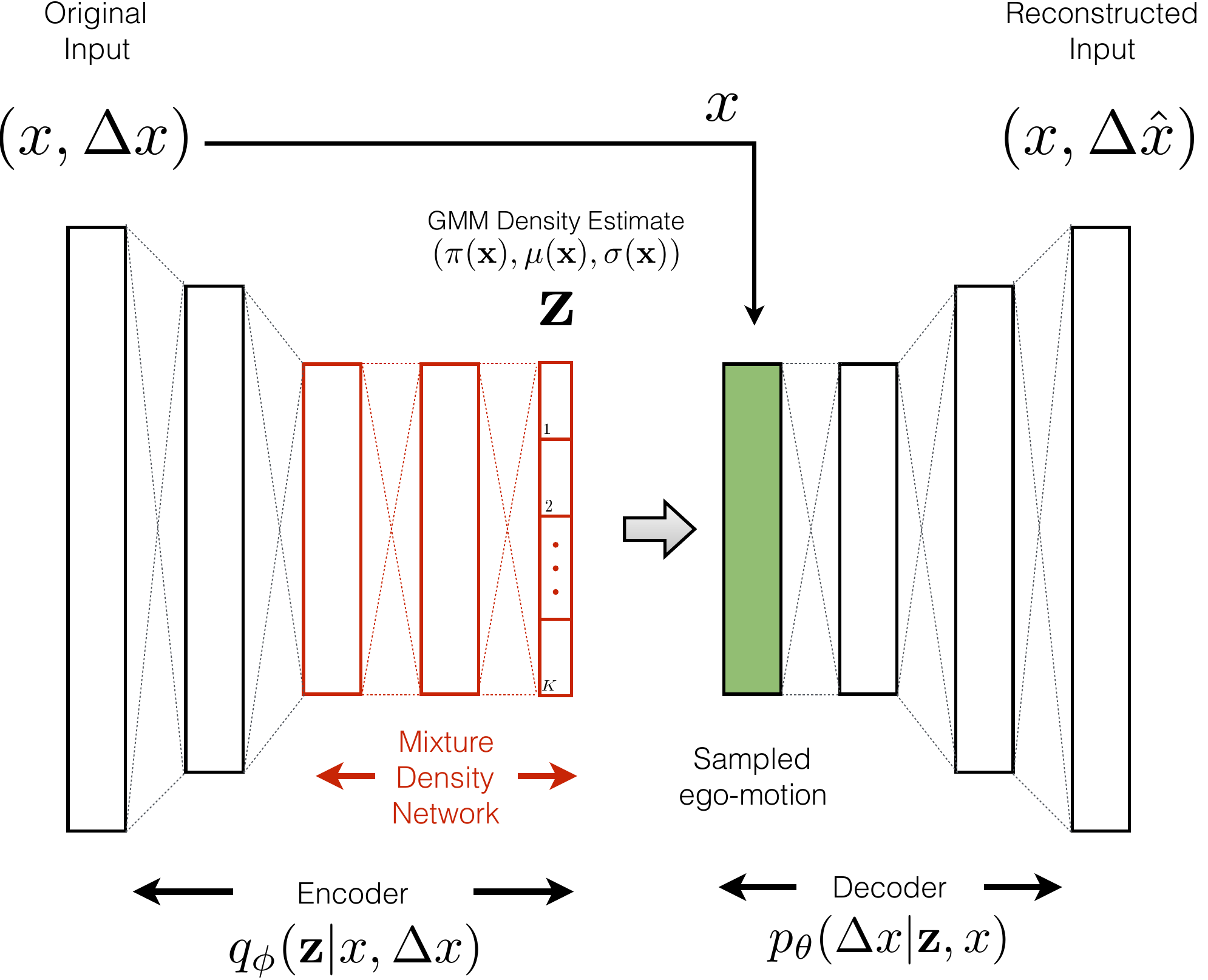}
  \caption{\textbf{Visual Ego-motion Learning Architecture: } We
    propose a visual ego-motion learning architecture that maps optical
    flow vectors (derived from feature tracking in an image sequence) to
    an ego-motion density estimate via a Mixture Density Network (MDN). By
    modeling the architecture as a Conditional Variational Autoencoder
    (C-VAE), our model is able to provide introspective reasoning and
    prediction for scene-flow conditioned on the ego-motion estimate and
    input feature location.}
  \label{fig:egomotion-architecture}
  \vspace{-5mm}
\end{figure}

With increasing levels of model specification for each domain, we
expect these algorithms to perform differently from others while
maintaining lesser generality across various optics and camera
configurations. Moreover, the strong dependence of these algorithms on
their model specification limits the ability to actively monitor and
optimize their intrinsic and extrinsic model parameters in an online
fashion. In addition to these concerns, autonomous systems today use
several sensors with varied intrinsic and extrinsic properties that
make system characterization tedious. Furthermore, these
algorithms and their parameters are fine-tuned on specific datasets
while enforcing little guarantees on their generalization performance
on new data.

To this end, we propose a fully trainable architecture for visual
odometry estimation in generic cameras with varied camera optics
(\textit{pinhole}, \textit{fisheye} and \textit{catadioptric}
lenses). In this work, we take a geometric approach by posing the
regression task of ego-motion as a density estimation problem. By
tracking salient features in the image induced by the ego-motion (via
Kanade-Lucas-Tomasi/KLT feature tracking), we learn the mapping from
these tracked flow features to a probability mass over the range of
likely ego-motion. We make the following contributions:
\begin{itemize}
  \item \textbf{A fully trainable ego-motion estimator}: We
introduce a fully-differentiable density estimation model for visual
ego-motion estimation that robustly captures the inherent ambiguity
and uncertainty in relative camera pose estimation (See
Figure~\ref{fig:egomotion-architecture}).
\item \textbf{Ego-motion for generic camera optics}: Without imposing
any constraints on the type of camera optics, we propose an approach
that is able to recover ego-motions for a variety of camera models
including \textit{pinhole}, \textit{fisheye} and \textit{catadioptric}
lenses.
\item\textbf{Bootstrapped ego-motion training and refinement}: We propose a
bootstrapping mechanism for autonomous systems whereby a robot
self-supervises the ego-motion regression task. By fusing information
from other sensor sources including GPS and INS (Inertial Navigation
Systems), these indirectly inferred trajectory estimates serve as
ground truth target poses/outputs for the aforementioned regression
task. Any newly introduced camera sensor can now leverage this
information to learn to provide visual ego-motion estimates without
relying on an externally provided ground truth source.
\item\textbf{Introspective reasoning via scene-flow predictions}: We
develop a generative model for optical flow prediction that can be
utilized to perform outlier-rejection and scene flow reasoning. 
\end{itemize}
Through experiments, we provide a thorough analysis of ego-motion
recovery from a variety of camera models including pinhole, fisheye
and catadioptric cameras. We expect our general-purpose approach to be
robust, and easily tunable for accuracy during
operation. We illustrate the robustness and generality of our approach
and provide our findings in Section~\ref{sec:experiments}.




\section{Related Work}\label{sec:related-work} Recovering relative
camera poses from a set of images is a well studied problem under the
context of Structure-from-Motion
(SfM)~\cite{triggs1999bundle,hartley2003multiple}. SfM is usually
treated as a non-linear optimization problem, where the camera poses
(extrinsics), camera model parameters (intrinsics), and the 3D scene
structure are jointly optimized via non-linear
least-squares~\cite{triggs1999bundle}.


\textbf{Unconstrained VO}: Visual odometry, unlike incremental
Structure-from-Motion, only focuses on determining the 3D camera pose
from sequential images or video imagery observed by a monocular
camera. Most of the early work in VO was done primarily to determine
vehicle
egomotion~\cite{moravec1980obstacle,matthies1989dynamic,olson2000robust}
in 6-DOF, especially in the Mars planetary rover. Over the years
several variants of the VO algorithm were proposed, leading up to the
work of Nister et al.~\cite{nister2004visual}, where the authors
proposed the first real-time and scalable VO algorithm. In their work,
they developed a 5-point minimal solver coupled with a RANSAC-based
outlier rejection scheme~\cite{fischler1981random} that is still
extensively used today. Other researchers~\cite{corke2004omnidirectional}
have extended this work to various camera types including catadioptric
and fisheye lenses.

\textbf{Constrained VO}: While the classical VO objective does not
impose any constraints regarding the underlying motion manifold or
camera model, it however contains several failure modes that make it
especially difficult to ensure robust operation under arbitrary scene
and lighting conditions. As a result, imposing egomotion constraints has
been shown to considerably improve accuracy, robustness, and run-time
performance. One particularly popular strategy for VO estimation in
vehicles is to enforce planar homographies during matching features
on the ground
plane~\cite{liang2002visual,ke2003transforming},
thereby being able to robustly recover both relative orientation and
absolute scale. For example, Scaramuzza et
al.~\cite{scaramuzza20111,scaramuzza2009real} introduced a novel
1-point solver by imposing the vehicle's non-holonomic motion
constraints, thereby speeding up the VO estimation up to 400Hz. 


\textbf{Data-driven VO}: While several model-based methods have been
developed specifically for the VO problem, a few have attempted to
solve it with a data-driven approach. Typical approaches have
leveraged dimensionality reduction techniques by learning a
reduced-dimensional subspace of the optical flow vectors induced by
the
egomotion~\cite{roberts2009learning}. In~\cite{ciarfuglia2014evaluation},
Ciarfuglia et al. employ Support Vector Regression (SVR) to recover
vehicle egomotion (3-DOF). The authors further build upon their
previous result by swapping out the SVR module with an end-to-end
trainable convolutional neural network~\cite{costante2016exploring}
while showing improvements in the overall performance on
the KITTI odometry benchmark~\cite{Geiger2012CVPR}. Recently, Clarke
et al.~\cite{wen2016vinet} introduced a visual-inertial odometry
solution that takes advantage of a neural-network architecture to learn
a mapping from raw inertial measurements and sequential imagery to
6-DOF pose estimates.  By posing visual-inertial odometry (VIO) as a
sequence-to-sequence learning problem, they developed a neural network
architecture that combined convolutional neural networks with Long
Short-Term Units (LSTMs) to fuse the independent sensor measurements
into a reliable 6-DOF pose estimate for ego-motion. Our work closely
relates to these data-driven approaches that have recently been
developed. We provide a qualitative comparison of how our approach is
positioned within the visual ego-motion estimation landscape in
Table~\ref{table:vo-landscape}. 


\begin{table}[h]
  \centering
  \scriptsize
  \rowcolors{2}{gray!25}{white}
  {\renewcommand{\arraystretch}{1} 
    {\setlength{\tabcolsep}{0.2mm}
      \begin{tabular}{lM{1cm}M{1cm}M{1cm}M{1.25cm}}
        \toprule
        \centering
        \textbf{Method Type} & \textbf{Varied Optics} & \textbf{Model Free} & \textbf{Robust} 
        & \textbf{Self Supervised} \\ \midrule
        \textit{Traditional VO~\cite{scaramuzza2011visual}}
                        & $\xmark$ & $\xmark$ & $\cmark$ & $\xmark$ \\ 
        \textit{End-to-end VO~\cite{costante2016exploring,wen2016vinet}}
                        & $\xmark$ & $\cmark$ & $\cmark$ & $\xmark$ \\ 
        \textit{This work}
                        & $\cmark$ & $\cmark$ & $\cmark$ & $\cmark$ \\
        \bottomrule
      \end{tabular}}}
\caption{\textbf{Visual odometry landscape}: A qualitative comparison of how our approach is
  positioned amongst existing solutions to ego-motion
  estimation.}
\label{table:vo-landscape}\vspace{-5mm}
\end{table}




\section{Ego-motion regression}\label{sec:procedure}

As with most ego-motion estimation solutions, it is imperative to
determine the minimal parameterization of the underlying motion
manifold. In certain restricted scene structures or motion manifolds,
several variants of ego-motion estimation are
proposed~\cite{scaramuzza20111,liang2002visual,ke2003transforming,scaramuzza2009real}.
However, we consider the case of modeling cameras with varied optics
and hence are interested in determining the full range of ego-motion,
often restricted, that induces the pixel-level optical flow. This
allows the freedom to model various unconstrained and partially
constrained motions that typically affect the overall robustness of
existing ego-motion algorithms.  While model-based approaches have
shown tremendous progress in accuracy, robustness, and run-time
performance, a few recent data-driven approaches have been shown to
produce equally compelling
results~\cite{costante2016exploring,wen2016vinet,konda2015learning}. An
adaptive and trainable solution for relative pose estimation or
ego-motion can be especially advantageous for several reasons: (i) a
general-purpose end-to-end trainable model architecture that applies to a
variety of camera optics including pinhole, fisheye, and catadioptric
lenses; (ii) simultaneous and continuous optimization over both ego-motion estimation
and camera parameters (intrinsics and extrinsics that are implicitly
modeled); and (iii) joint reasoning over resource-aware
computation and accuracy within the same architecture is amenable.  We
envision that such an approach is especially beneficial in the context
of bootstrapped (or weakly-supervised) learning in robots, where the
supervision in ego-motion estimation for a particular camera can be
obtained from the fusion of measurements from other robot
sensors (GPS, wheel encoders etc.).


Our approach is motivated by previous minimally parameterized
models~\cite{scaramuzza20111,scaramuzza2009real} that are able to
recover ego-motion from a \textit{single tracked feature}. We find
this representation especially appealing due to the simplicity and
flexibility in~\textit{pixel-level} computation. Despite the reduced
complexity of the input space for the mapping problem, recovering the
full 6-DOF ego-motion is ill-posed due to the inherently
under-constrained system. However, it has been previously shown that
under non-holonomic vehicle motion, camera ego-motion may be fully
recoverable up to a sufficient degree of accuracy using a single
point~\cite{scaramuzza20111,scaramuzza2009real}.

We now focus on the specifics of
the ego-motion regression objective. Due to the under-constrained
nature of the prescribed regression problem, the pose estimation is
modeled as a density estimation problem over the range of possible
ego-motions\footnote{\scriptsize Although the parametrization is
  maintained as $SE(3)$, it is important to realize
that the nature of most autonomous car datasets involve a
lower-dimensional ($SE(2)$) motion manifold}, conditioned on the
input flow features. It is important to note that the output of the
proposed model is a density estimate
$p(\hat{\Bz}_{t-1,t}\vert\Bx_{t-1,t})$ for every feature tracked
between subsequent frames. 



\subsection{Density estimation for
  ego-motion}\label{subsec:density-egomotion}
In typical associative mapping problems, the joint probability density
$p(\Bx,\Bz)$ is decomposed into the product of two terms: (i) $p(\Bz
\vert \Bx)$: the conditional density of the target pose $\Bz \in
SE(3)$ conditioned on the input feature correspondence $\Bx = (\vx,
\Delta\vx)$ obtained from sparse optical flow (KLT)~\cite{birchfield2007klt} (ii)
$p(\Bx)$: the unconditional density of the input data $\Bx$. While we
are particularly interested in the first term $p(\Bz\vert\Bx)$ that
predicts the range of possible values for $\Bz$ given new values of
$\Bx$, we can observe that the density $p(\Bx) = \sum_z p(\Bx,\Bz)
d\Bz$ provides a measure of how well the prediction is captured by the
trained model.

The critical component in estimating the ego-motion belief is the
ability to accurately predict the conditional probability distribution
$p(\Bz\vert\Bx)$ of the pose estimates that is induced by the given
input feature $\vx$ and the flow $\Delta\vx$. Due to its powerful and
rich modeling capabilities, we use a \textit{Mixture Density Network}
(MDN)~\cite{bishop1994mixture} to parametrize the conditional density
estimate. MDNs are a class of end-to-end trainable
(fully-differentiable) density estimation techniques that leverage
conventional neural networks to regress the parameters of a generative
model such as a finite Gaussian Mixture Model (GMM). The powerful representational
capacity of neural networks coupled with rich probabilistic modeling
that GMMs admit, allows us to model multi-valued or multi-modal
beliefs that typically arise in inverse problems such as visual
ego-motion.

For each of the $F$ input flow features $\Bx_i$ extracted via KLT, the
conditional probability density of the target pose data $\Bz_i$
(Eqn~\ref{eq:cdf}) is
represented as a convex combination of $K$ Gaussian components,
\vspace{-1mm}
\begin{align}\label{eq:cdf}
  p(\Bz_i \mid \Bx_i) = \sum_{k=1}^{K} \pi_k(\Bx_i) \mathcal{N}(\Bz \mid \mu_k(\Bx_i), \sigma_{k}^2(\Bx_i))
\end{align}where $\pi_k(\Bx)$ is the mixing coefficient for the $k$-th component
as specified in a typical GMM. The Gaussian
kernels are parameterized by their mean vector $\mu_k(\Bx)$ and
diagonal covariance $\sigma_{k}(\Bx)$. It is important to note that
the parameters $\pi_k(\Bx), \mu_k(\Bx)$, and $\sigma_k(\Bx)$ are
general and continuous functions of $\Bx$. This allows us to model
these parameters as the output ($a^{\pi}$, $a^{\mu}$,
$a^{\sigma}$) of a conventional neural network which takes $\Bx$ as
its input. Following~\cite{bishop1994mixture}, the outputs of the
neural network are constrained as follows: (i) The mixing coefficients
must sum to 1, i.e. $\sum_{K}\pi_k(\Bx) = 1$ where $0 \leq \pi_k(\Bx)
\leq 1$. This is accomplished via the \textit{softmax} activation as
seen in Eqn~\ref{eq:mdn-pi}. (ii) Variances $\sigma_k(\Bx)$ are
strictly positive via the \textit{exponential} activation
(Eqn~\ref{eq:mdn-sigma}).
\begin{align}
  &\pi_k(\Bx) = \frac{\exp(a_k^\pi)} { \sum_{l=1}^{K} \exp(a_l^\pi) }\label{eq:mdn-pi} \\
  &\sigma_k(\Bx) = \exp(a_k^\sigma), \hspace{4mm}
    \mu_{k}(\Bx) = a_{k}^\mu \label{eq:mdn-sigma} \vspace{4mm}\\
  \mathcal{L_{MDN}} = -
  &\sum_{n=1}^{N} \ln \Bigg\{ \sum_{k=1}^{K} \pi_k(\Bx_n) \mathcal{N}(\Bz \mid
    \mu_k(\Bx_n), \sigma_{k}^2(\Bx_n)) \Bigg\} \label{eq:nll}
\end{align}
The proposed model is learned end-to-end by maximizing the data
log-likelihood, or alternatively minimizing the negative log-likelihood
(denoted as $\mathcal{L_{MDN}}$ in Eqn~\ref{eq:nll}), given the $F$ input feature tracks ($\Bx_1\dots\Bx_F$) and expected
ego-motion estimate $\Bz$. The resulting ego-motion density estimates
$p(\Bz_i\vert\Bx_i)$
obtained from
each individual flow vectors $\Bx_i$ are then fused by taking the
product of their densities. However, to
maintain tractability of density products, only the mean and
covariance corresponding to the largest mixture coefficient (i.e. most
likely mixture mode) for each feature is considered for subsequent trajectory
optimization (See Eqn~\ref{eq:density-products}).
\begin{align}\label{eq:density-products}
  p(\Bz \vert \Bx) \simeq \prod_{i=1}^{F} \max_{k} \Big\{\pi_k(\Bx_i)
  \mathcal{N}(\Bz_i \mid \mu_k(\Bx_i), \sigma_{k}^2(\Bx_i)) \Big\}
\end{align}


%
\begin{figure}[!t]
  \centering
  \includegraphics[width=\columnwidth]{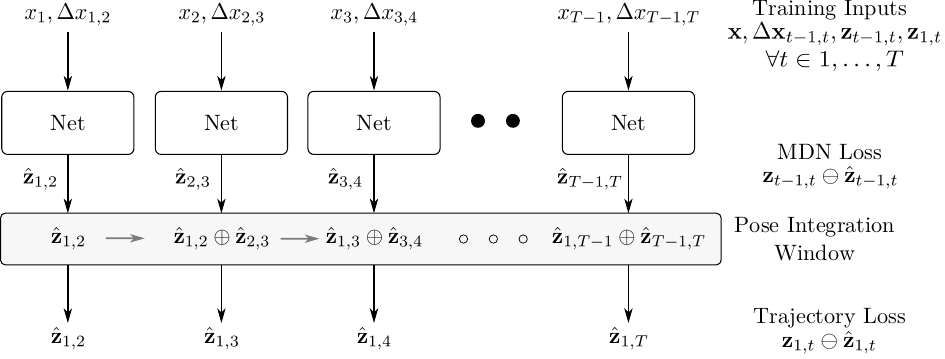}
  \caption{\textbf{Windowed trajectory optimization}: An illustration
    of the losses introduced for training frame-to-frame ego-motion
    (\textit{local}) and windowed ego-motion (\textit{global}) by
    compounding the poses determined from each of the individual
    frame-to-frame measurements.}
  \label{fig:losses-illustration}
  \vspace{-6mm}
\end{figure}

\subsection{Trajectory
  optimization}\label{sec:trajectory-optimization} While minimizing the MDN loss
($\mathcal{L}_{MDN}$) as described above provides a reasonable
regressor for ego-motion estimation, it is evident that optimizing
frame-to-frame measurements do not ensure long-term consistencies in
the ego-motion trajectories obtained by integrating these regressed
estimates. As one expects, the integrated trajectories are sensitive
to even negligible biases in the ego-motion regressor.

\textbf{Two-stage optimization}: To circumvent the aforementioned issue, we introduce a
second optimization stage that jointly minimizes the 
\textit{local} objective ($\mathcal{L}_{MDN}$) with a \textit{global} objective that
minimizes the error incurred between the overall trajectory and the
trajectory obtained by integrating the regressed pose estimates
obtained via the \textit{local} optimization. This allows the
\textit{global} optimization stage to have a warm-start with an almost
correct initial guess for the network parameters.

\begin{figure}[!t]
  \centering
  \centering 
  {\renewcommand{\arraystretch}{0.4} 
    {\setlength{\tabcolsep}{0.2mm}
      \begin{tabular}{cccc}
        \includegraphics[width=0.25\columnwidth]{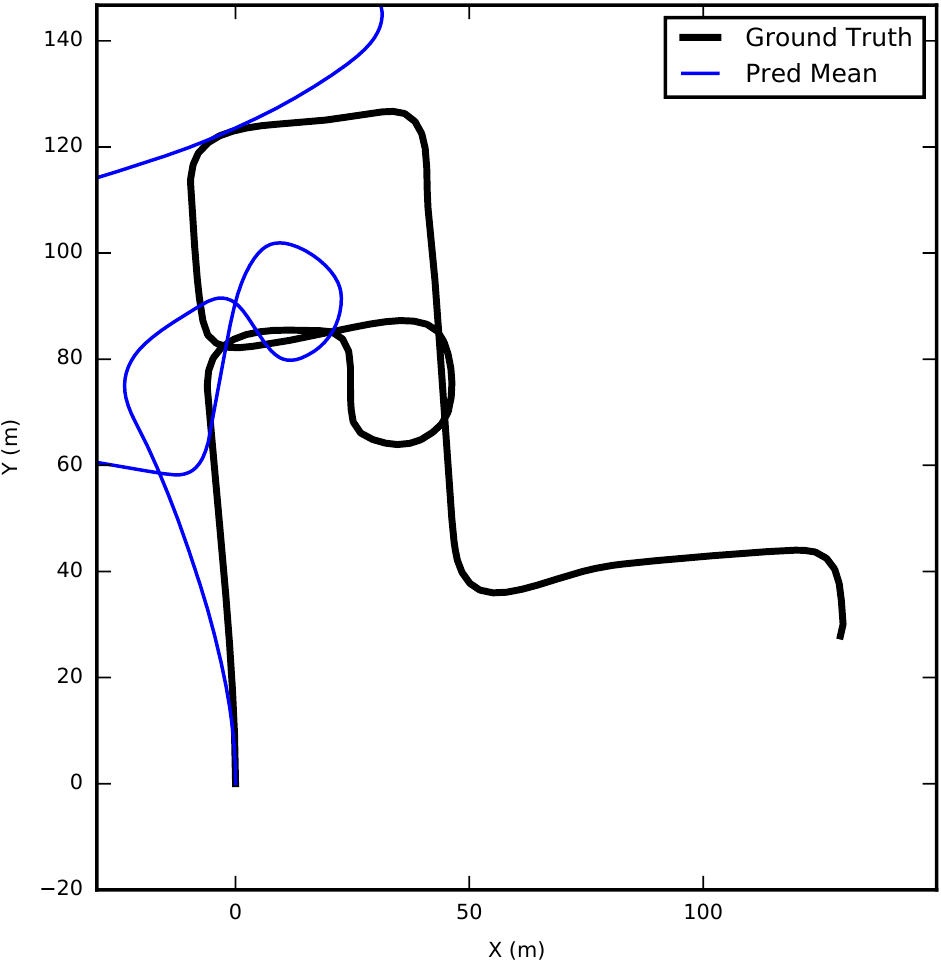} 
        &\includegraphics[width=0.25\columnwidth]{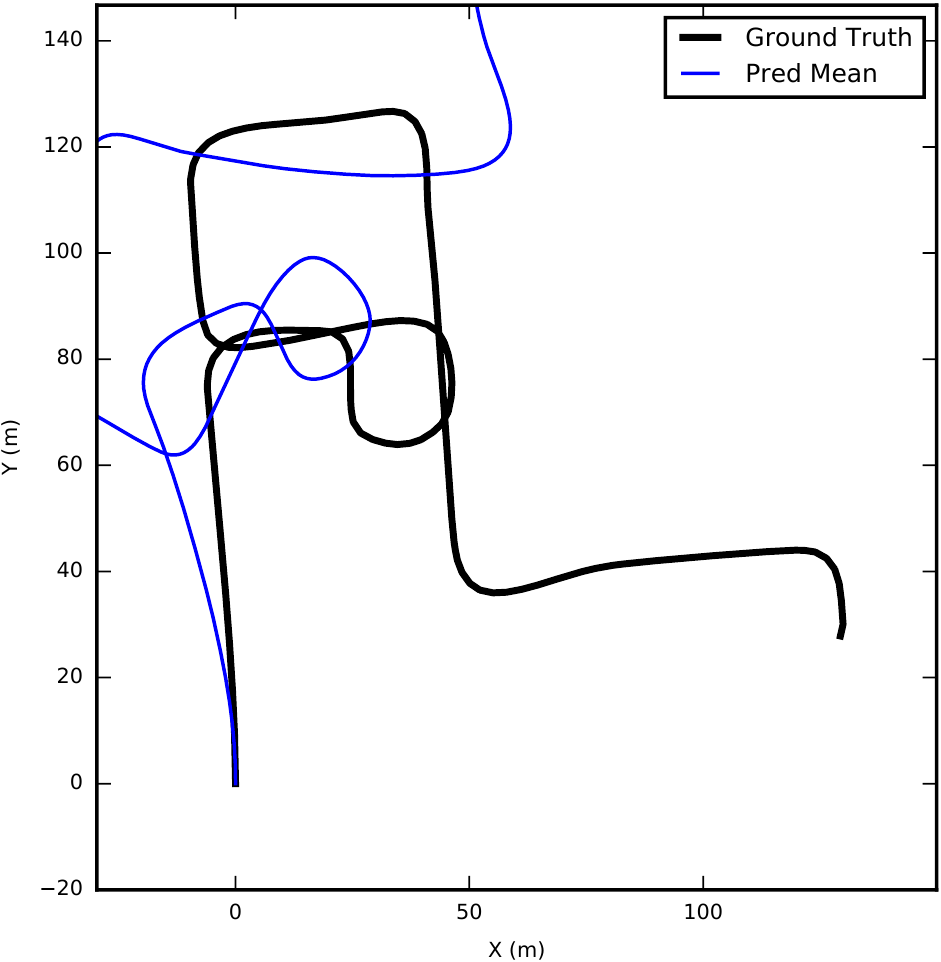}
        &\includegraphics[width=0.25\columnwidth]{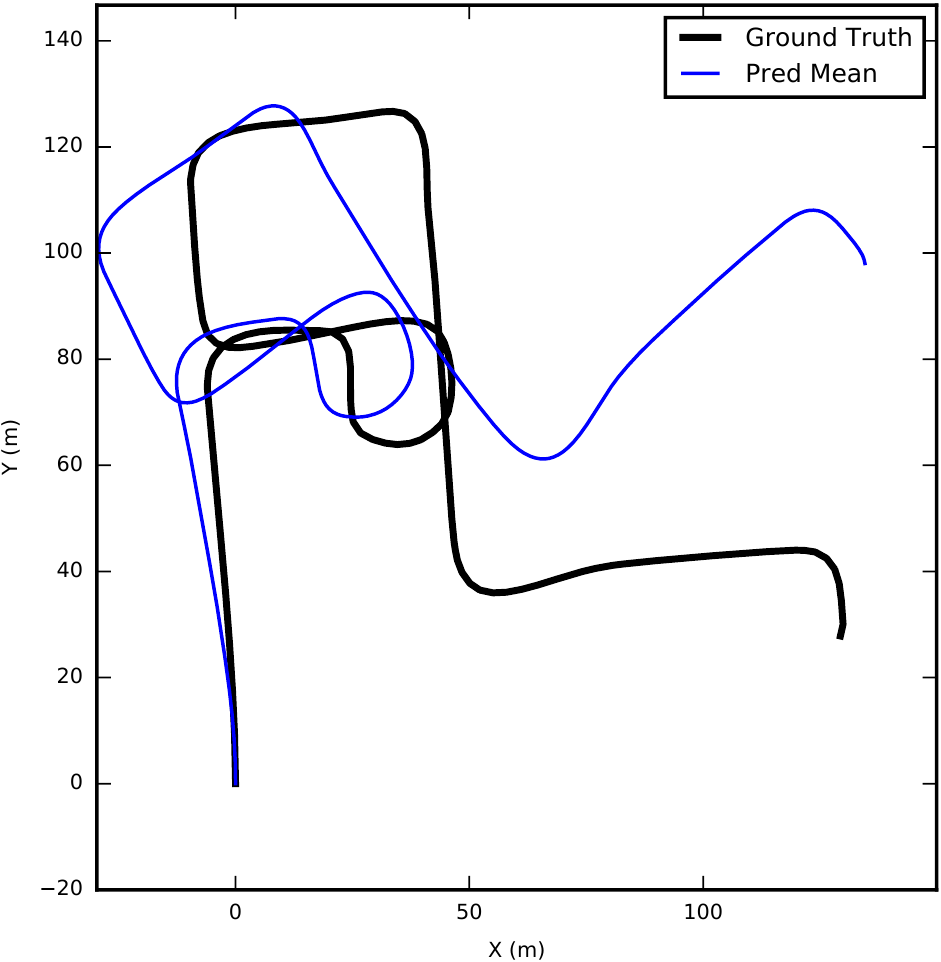}
        &\includegraphics[width=0.25\columnwidth]{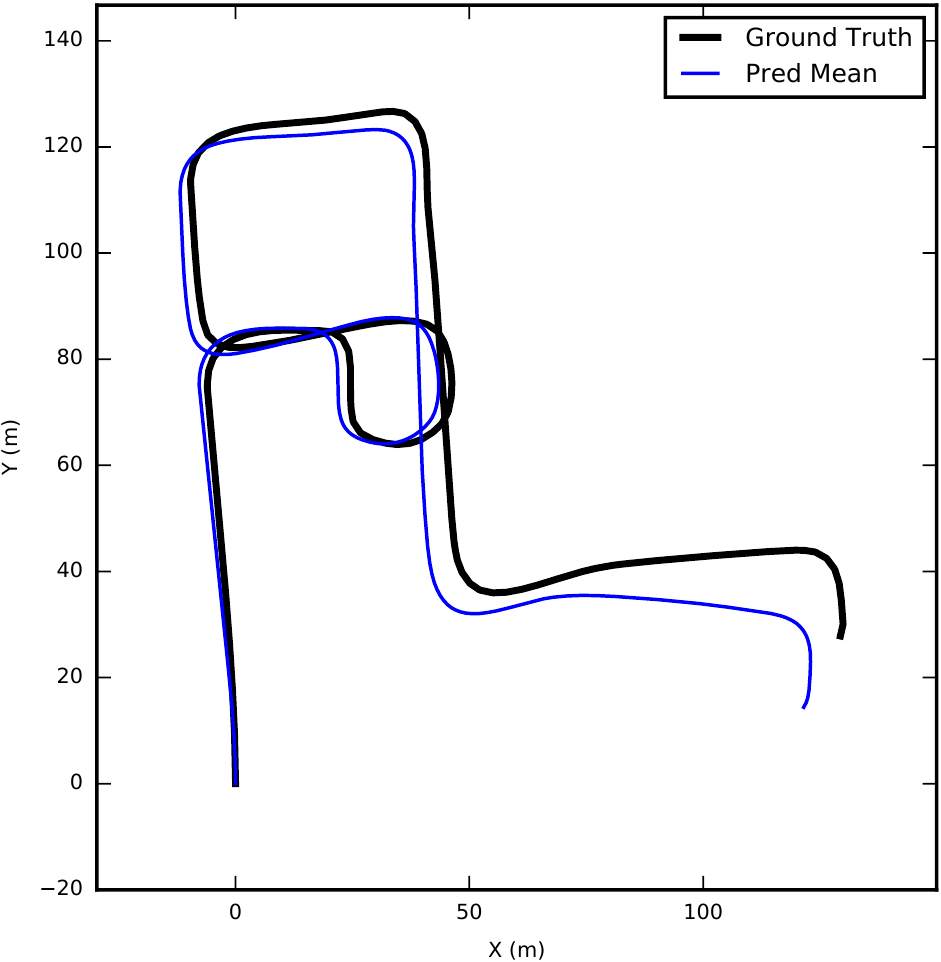}\\
        {\scriptsize \textbf{Stage 1}} & {\scriptsize \textbf{Stage 2}}
        &{\scriptsize \textbf{Stage 2}} & {\scriptsize \textbf{Stage 2}}\\
        {\scriptsize (Final)} & {\scriptsize (Epoch 4)}
        &{\scriptsize (Epoch 8)} & {\scriptsize (Epoch 18)}
      \end{tabular}}}
  \caption{\textbf{Two-stage Optimization}: An illustration of the
    two-stage optimization procedure. The \textit{first} column shows
    the final solution after the first stage. Despite the
    minimization, the integrated trajectory is clearly biased and
    poorly matches the expected result. The \textit{second},
    \textit{third} and \textit{fourth} column shows the gradual improvement of the
    second stage (global
    minimization) and matches the expected ground truth trajectory
    better (i.e. estimates the regressor biases better).}
  \label{fig:two-stage-illustration}
  \vspace{-5mm}
\end{figure}

As seen in Eqn~\ref{eq:losses}, $\mathcal{L}_{TRAJ}$ pertains to the
overall trajectory error incurred by integrating the individual
regressed estimates over a batched window (we typically consider 200
to 1000 frames). This allows us to fine-tune the regressor to predict
valid estimates that integrate towards accurate long-term ego-motion
trajectories. As
expected, the model is able to roughly learn the curved trajectory
path, however, it is not able to make accurate predictions when
integrated for longer time-windows (due to the lack of the
\textit{global} objective loss term in Stage
1). Figure~\ref{fig:losses-illustration} provides a high-level
overview of the input-output relationships of the training procedure,
including the various network losses incorporated in the ego-motion
encoder/regressor. For illustrative purposes only, we refer the reader to
Figure~\ref{fig:two-stage-illustration} where we validate this
two-stage approach over a simulated dataset~\cite{Zhang2016ICRA}. 

In Eqn~\ref{eq:losses}, $\hat{\Bz}_{{\scriptsize t-1,t}}$ is the
frame-to-frame ego-motion estimate and the regression target/output of the MDN
function $F$, where $F:\Bx \mapsto \Big(\mu({\Bx}_{{\scriptsize
t-1,t}}), \sigma(\Bx_{{\scriptsize t-1,t}}), \pi(\Bx_{{\scriptsize
t-1,t}})\Big)$. $\hat{\Bz}_{1,t}$ is the overall trajectory
predicted by integrating the individually regressed frame-to-frame
ego-motion estimates and is defined by $\hat{\Bz}_{1,t} = \hat{\Bz}_{1,2} \oplus \hat{\Bz}_{2,3} \oplus
\dots \oplus \hat{\Bz}_{t-1,t}$.
\begin{equation}
  \begin{aligned}
    \mathcal{L_{\text{ENC}}} =
    \underbrace{\sum_{t} \mathcal{L}^t_{{\scriptsize MDN}}\Big( F(\Bx),
    \Bz_{{\scriptsize t-1,t}}\Big)}_{\text{MDN Loss}} + 
    \underbrace{\sum_{t} \mathcal{L}^t_{{\scriptsize TRAJ}}(\Bz_{1,t} \ominus
      \hat{\Bz}_{1,t})}_{\text{Overall Trajectory Loss}}
  \end{aligned}
  \label{eq:losses}           
\end{equation}
\vspace{-5mm}

%






\subsection{Bootstrapped learning for ego-motion
estimation}\label{sec:proc-bootstrapped} 
\begin{figure}[!b]
  \centering
  \includegraphics[width=\columnwidth]{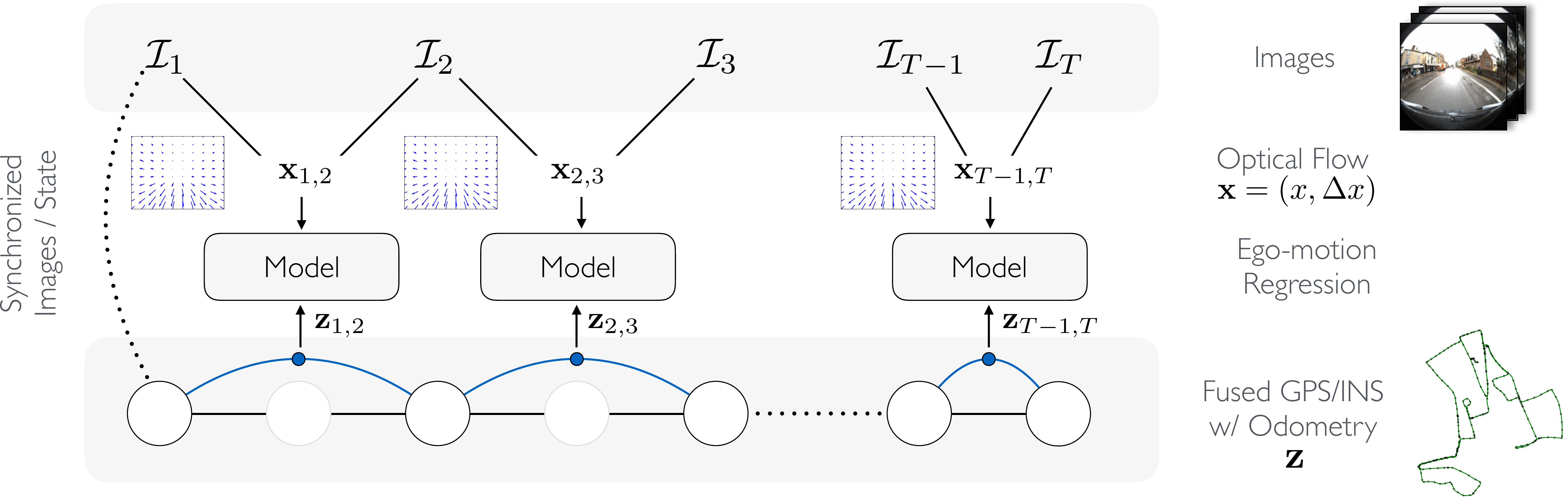}
  \caption{\textbf{Bootstrapped Ego-motion Regression}:
    Illustration of the bootstrap mechanism whereby a robot
    self-supervises the proposed ego-motion regression task in a new
    camera sensor by fusing information from other sensor sources
    such as GPS and INS.}
  \label{fig:egomotion-regression-illustration}
  \vspace{-4mm}
\end{figure}

Typical robot navigation
systems consider the fusion of visual odometry estimates with other
modalities including estimates derived from wheel encoders, IMUs, GPS
etc. Considering odometry estimates (for e.g. from wheel encoders)
as-is, the uncertainties in open-loop chains grow in
an unbounded manner. Furthermore, relative pose estimation may also be
inherently biased due to calibration errors that eventually contribute
to the overall error incurred. GPS, despite being noise-ridden,
provides an absolute sensor reference measurement that is especially
complementary to the open-loop odometry chain maintained with odometry
estimates. The probabilistic fusion of these two relatively
uncorrelated measurement modalities allows us to recover a
sufficiently accurate trajectory estimate that can be directly used
as ground truth data $\Bz$ (in Figure~\ref{fig:egomotion-regression-illustration}) for our supervised regression problem.




%

%
\begin{figure}[!b]
  \centering
  \includegraphics[width=\columnwidth]{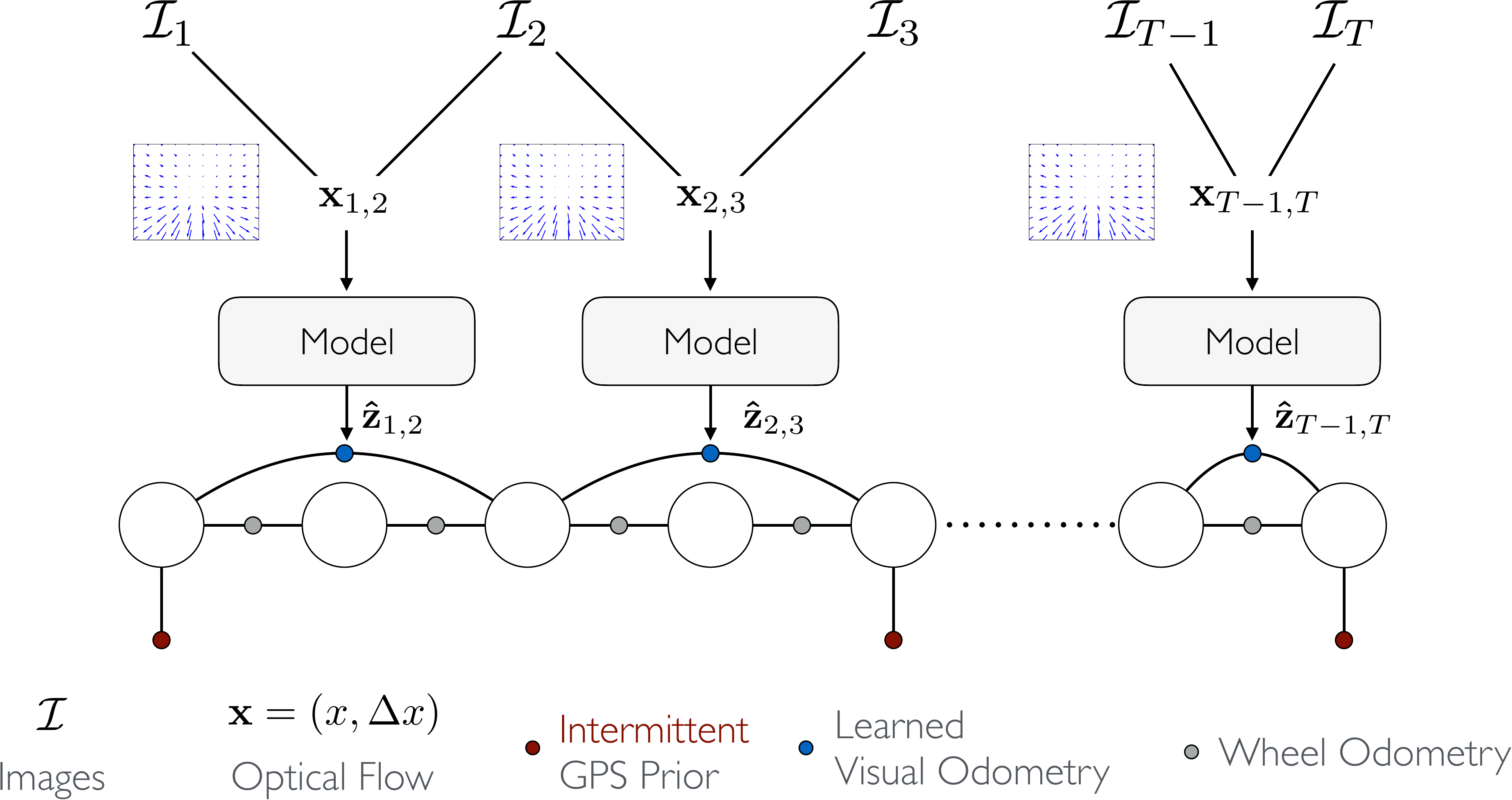}
  \caption{\textbf{Learned Ego-motion Deployment}:
    During model deployment, the learned visual-egomotion model
    provides valuable relative pose constraints to augment the standard
    navigation-based sensor fusion (GPS/INS and wheel encoder odometry
    fusion).}
  \label{fig:egomotion-deployment-illustration}
  \vspace{-2mm}
\end{figure}

The indirect recovery of training data from the fusion of other sensor
modalities in robots falls within the \textit{self-supervised or
bootstrapped} learning paradigm. We envision this capability to be
especially beneficial in the context of life-long learning in future
autonomous systems. Using the fused and optimized pose estimates $\Bz$
(recovered from GPS and odometry estimates), we are able to recover
the required input-output relationships for training visual ego-motion
for a completely new sensor (as illustrated in
Figure~\ref{fig:egomotion-regression-illustration}). Figure~\ref{fig:egomotion-deployment-illustration}
illustrates the realization of the learned model in a typical
autonomous system where it is treated as an additional sensor
source. Through experiments~\ref{sec:bootstrap-exp}, we illustrate
this concept with the recovery of ego-motion in a robot car equipped
with a GPS/INS unit and a single camera.



\subsection{Introspective Reasoning for Scene-Flow Prediction} Scene
flow is a fundamental capability that provides directly measurable
quantities for ego-motion analysis. The flow observed by sensors
mounted on vehicles is a function of the inherent scene depth, the
relative ego-motion undergone by the vehicle, and the intrinsic and
extrinsic properties of the camera used to capture it. As with any
measured quantity, one needs to deal with sensor-level noise
propagated through the model in order to provide robust
estimates. While the input flow features are an indication of
ego-motion, some of the features may be corrupted due to lack of or
ambiguous visual texture or due to flow induced by the dynamics of
objects other than the ego-motion itself. Evidently, we observe that
the dominant flow is generally induced by ego-motion itself, and it is
this flow that we intend to fully recover via a conditional
variational auto-encoder (C-VAE). By inverting the regression problem,
we develop a generative model able to predict the most-likely flow
$\hat{\Delta x}$ induced given an ego-motion estimate $\Bz$, and
feature location $x$. We propose a scene-flow specific autoencoder
that encodes the implicit egomotion observed by the sensor, while
jointly reasoning over the latent depth of each of the individual
tracked features.  


\begin{equation}
  \begin{aligned}
    \mathcal{L_{\text{CVAE}}} =& \mathbb{E}{\big [} \log p_{\theta}(\Delta x
    | \mathbf{z},x) {\big]}\\
                              &- D_{KL}\big[q_{\phi}(\mathbf{z} |
                              x,\Delta x) ||
                              p_{\theta}(\mathbf{z}|x)\big] 
  \end{aligned}
  \label{eq:cvae}                              
\end{equation}

Through the proposed denoising autoencoder model, we are also able to
attain an introspection mechanism for the presence of outliers. We
incorporate this additional module via an auxiliary loss as specified
in Eqn~\ref{eq:cvae}. An illustration of these flow
predictions are shown in Figure~\ref{fig:evaluation-flow-prediction}.


\section{Experiments}\label{sec:experiments} In this section, we
provide detailed experiments on the performance, robustness and
flexibility of our proposed approach on various datasets. Our approach
differentiates itself from existing solutions on various fronts as
shown in Table~\ref{table:vo-landscape}. We evaluate the performance
of our proposed approach on various publicly-available datasets
including the KITTI dataset~\cite{Geiger2012CVPR}, the Multi-FOV synthetic
dataset~\cite{Zhang2016ICRA} (pinhole, fisheye, and catadioptric
lenses), an omnidirectional-camera
dataset~\cite{schonbein2014omnidirectional}, and on the Oxford Robotcar
1000km Dataset~\cite{maddern20161}.

\begin{figure}[!t]
  \centering 
  {\renewcommand{\arraystretch}{0.4} 
    {\setlength{\tabcolsep}{0.4mm}
      \begin{tabular}{cccc}
        \rotatebox[x=-6mm]{90}{\scriptsize \textbf{Image}}& 
          \includegraphics[width=0.29\columnwidth,frame={\fboxrule}]{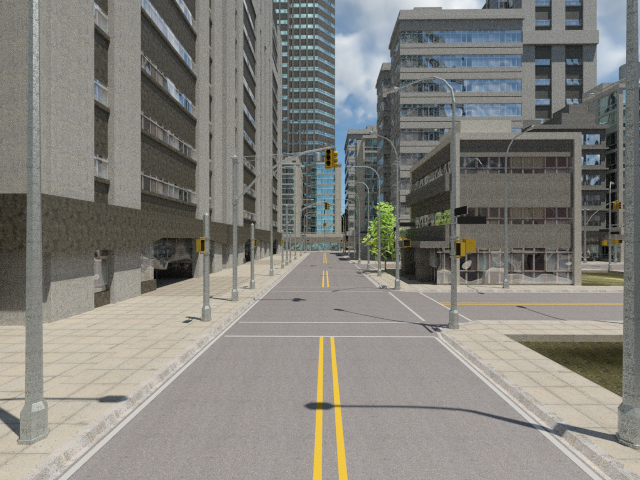}&
        \includegraphics[width=0.29\columnwidth,frame={\fboxrule}]{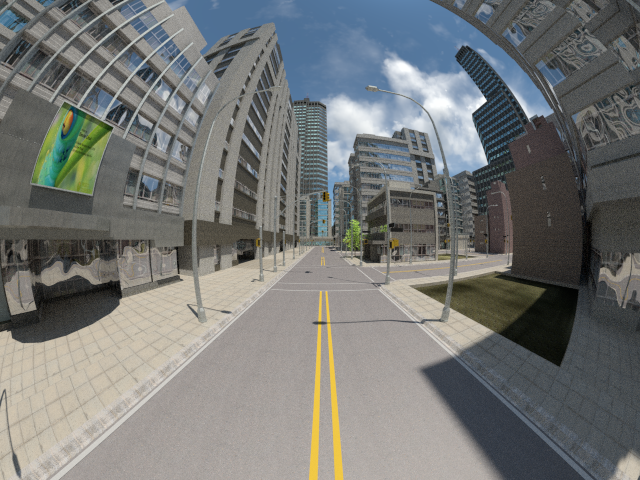}&
        \includegraphics[width=0.29\columnwidth,frame={\fboxrule}]{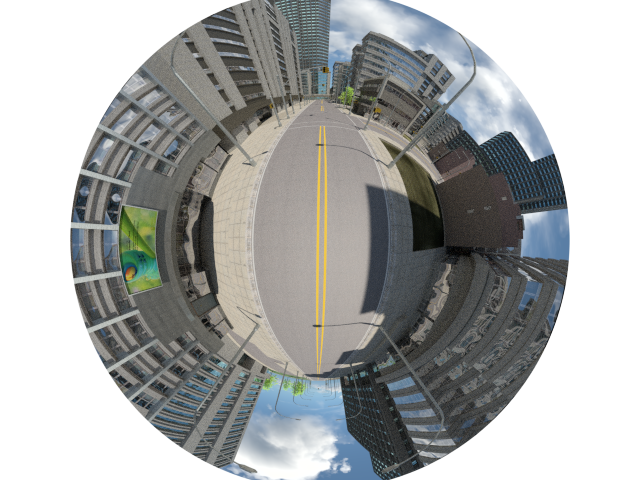}\\
        \rotatebox[x=-5.5mm]{90}{\scriptsize \textbf{Forward}}&
          \includegraphics[width=0.3\columnwidth]{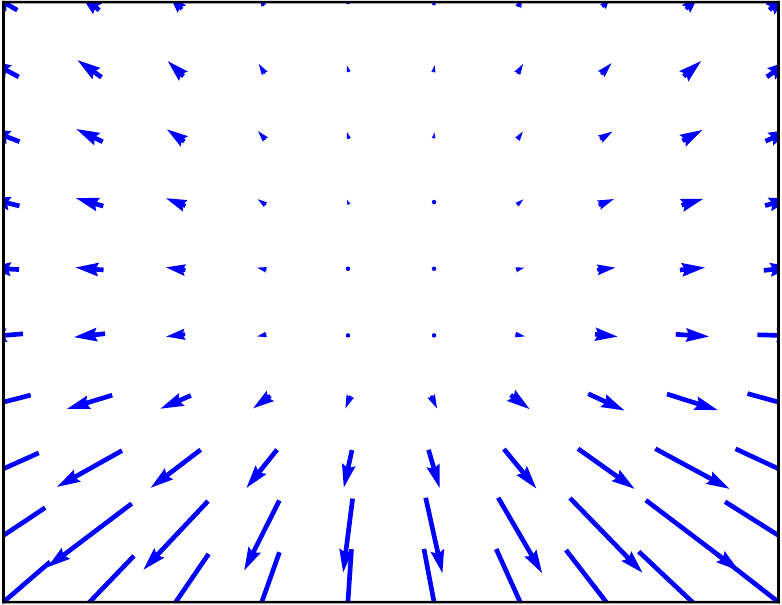}&
        \includegraphics[width=0.3\columnwidth]{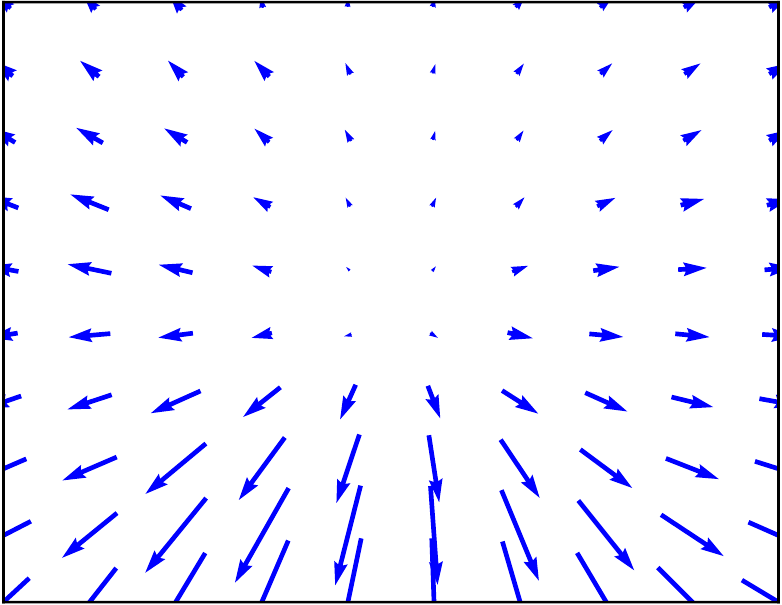}&
        \includegraphics[width=0.3\columnwidth]{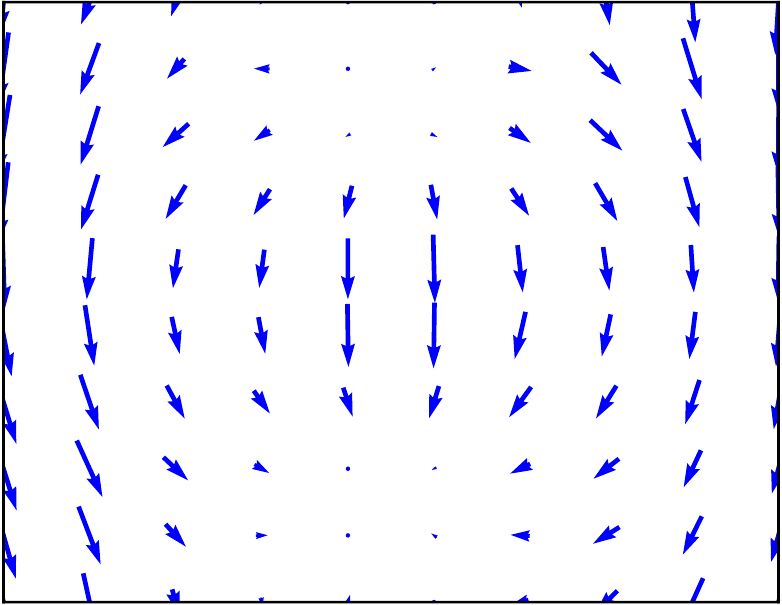}\\
        & \scriptsize \textbf{(a) Pinhole} & \scriptsize \textbf{(b) Fisheye} & \scriptsize \textbf{(c) Catadioptric}\\
      \end{tabular}}}
  \caption{\textbf{Introspective reasoning for scene-flow prediction}:
    Illustrated above are the dominant flow vectors corresponding to
    scene-flow given the corresponding ego-motion. While this module is
    not currently used in the ego-motion estimation, we expect it to be
    critical in outlier rejection. \textbf{Row 1}: Sample image from
    camera, \textbf{Row 2}: Flow induced by forward motion
  }
  \label{fig:evaluation-flow-prediction}
  \vspace{-6mm}
\end{figure}

Navigation solutions in autonomous systems today typically fuse
various modalities including GPS, odometry from wheel encoders and INS
to provide robust trajectory estimates over extended periods of
operation. We provide a similar solution by leveraging the learned
ego-motion capability described in this work, and fuse it with
intermittent GPS updates\footnote{\scriptsize For evaluation purposes
only, the absolute ground truth locations were added as weak priors on
datasets without GPS measurements}
(Secion~\ref{sec:performance}). While maintaining similar performance
capabilities (Table~\ref{tab:trajectory-errors}), we re-emphasize the benefits
of our approach over existing solutions:
\begin{itemize}
\item \textbf{Versatile}: With
a fully trainable model, our approach is able to
simultaneously reason over both ego-motion and implicitly modeled camera parameters
(\textit{intrinsics} and \textit{extrinsics}). Furthermore, online calibration and
parameter tuning is implicitly encoded within the same learning
framework.
\item \textbf{Model-free}: Without imposing any constraints
on the type of camera optics, our approach is able to recover
ego-motions for a variety of camera models including \textit{pinhole},
\textit{fisheye} and \textit{catadioptric}
lenses. (Section~\ref{sec:model-free})
\item \textbf{Bootstrapped training and refinement}: We illustrate a bootstrapped
learning example whereby a robot self-supervises the proposed
ego-motion regression task by fusing information from other sensor sources
including GPS and INS (Section~\ref{sec:bootstrap-exp})
\item \textbf{Introspective reasoning for scene-flow prediction}: Via the C-VAE
generative model,
we are able to reason/introspect over the predicted flow vectors in
the image given an ego-motion estimate. This provides an obvious
advantage in \textit{robust} outlier detection and identifying dynamic
objects whose flow vectors need to be disambiguated from the
ego-motion scene flow (Figure~\ref{fig:evaluation-flow-prediction})
\end{itemize}

%
\begin{figure*}[!t]
  \centering 
  {\renewcommand{\arraystretch}{0.6} 
    {\setlength{\tabcolsep}{0.2mm}
      \begin{tabular}{cccc}
        \includegraphics[height=1.55in]{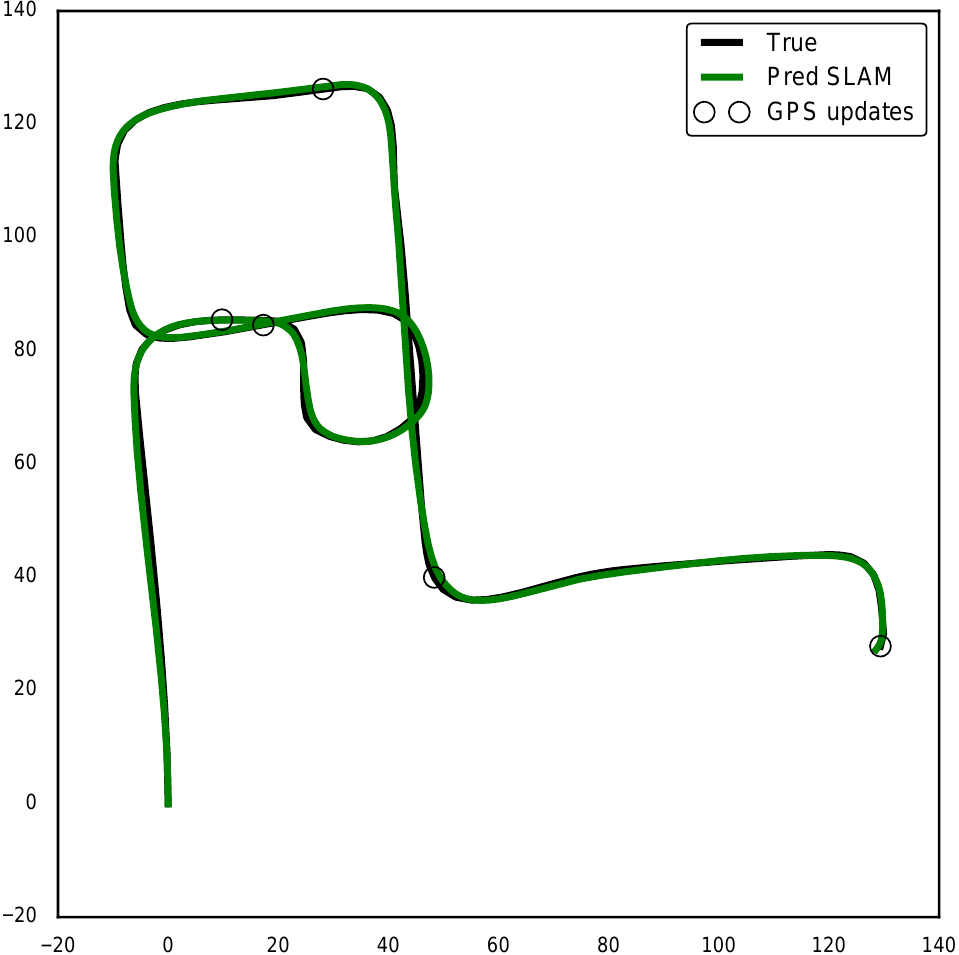}
        &\includegraphics[height=1.55in]{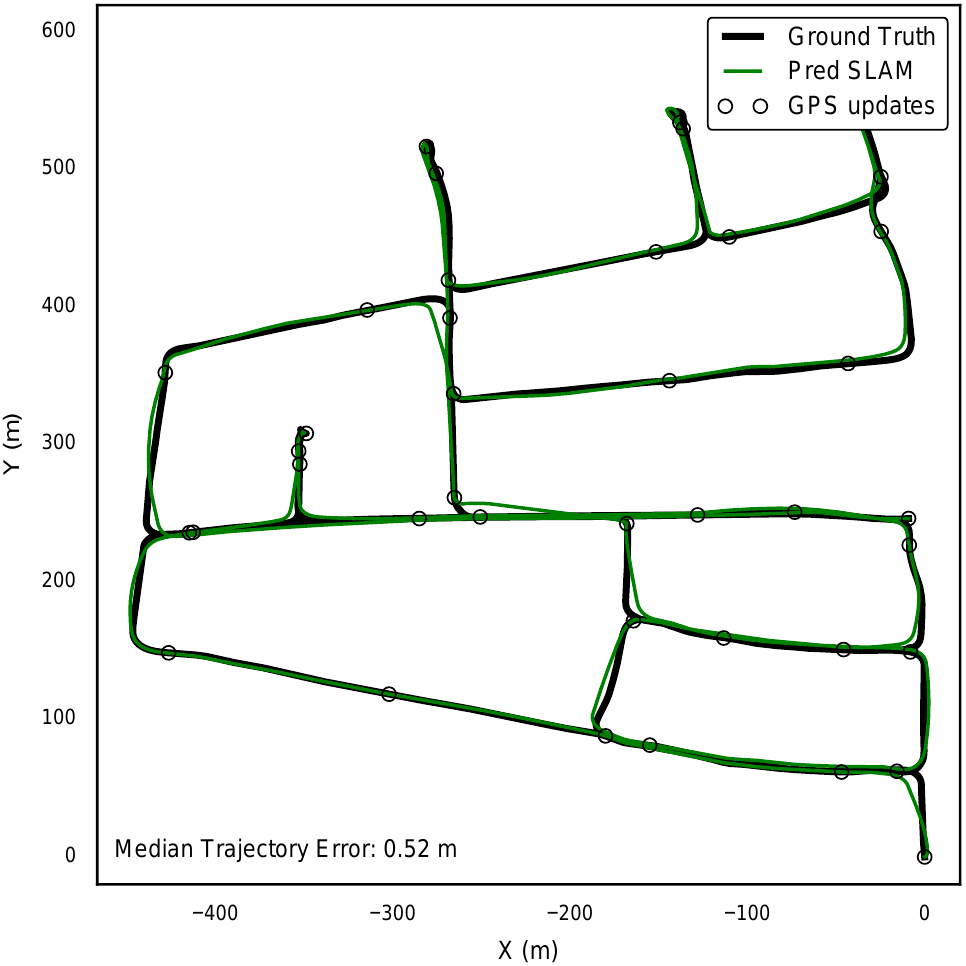}
        &\includegraphics[height=1.55in]{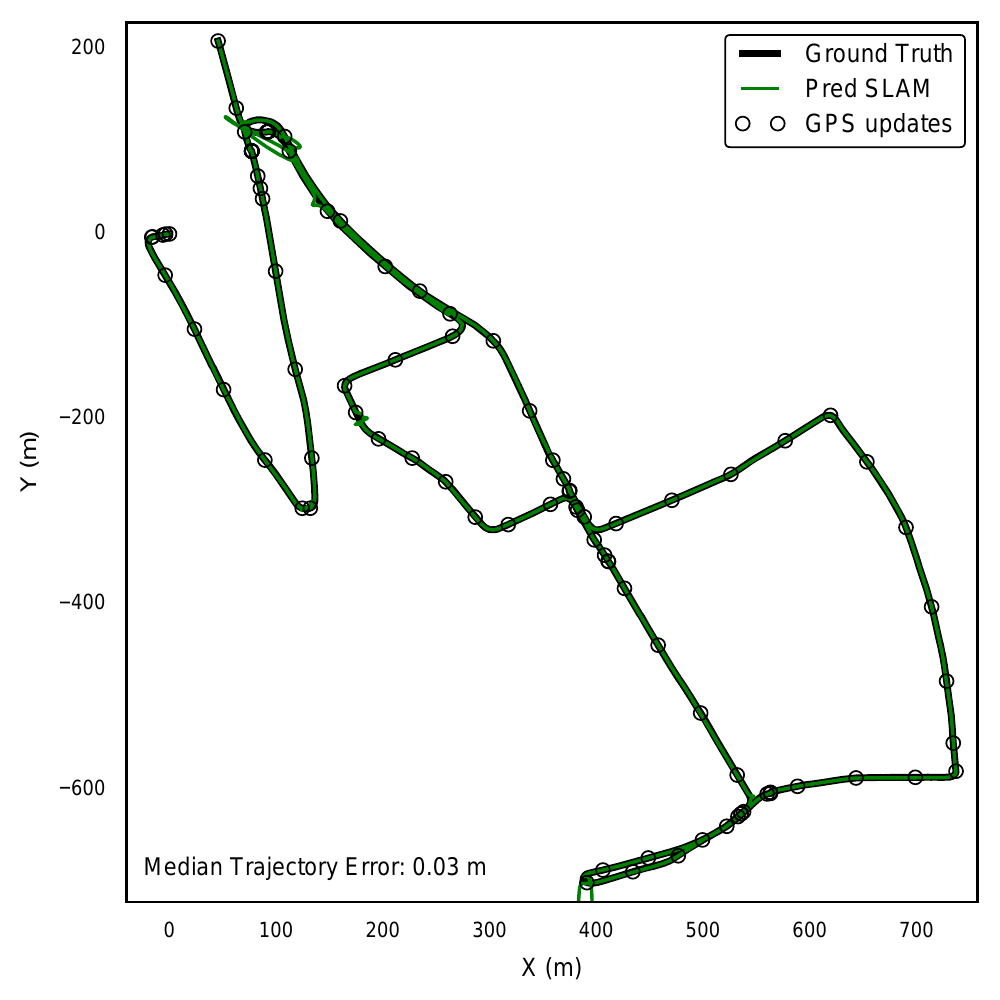}
        &\includegraphics[height=1.55in]{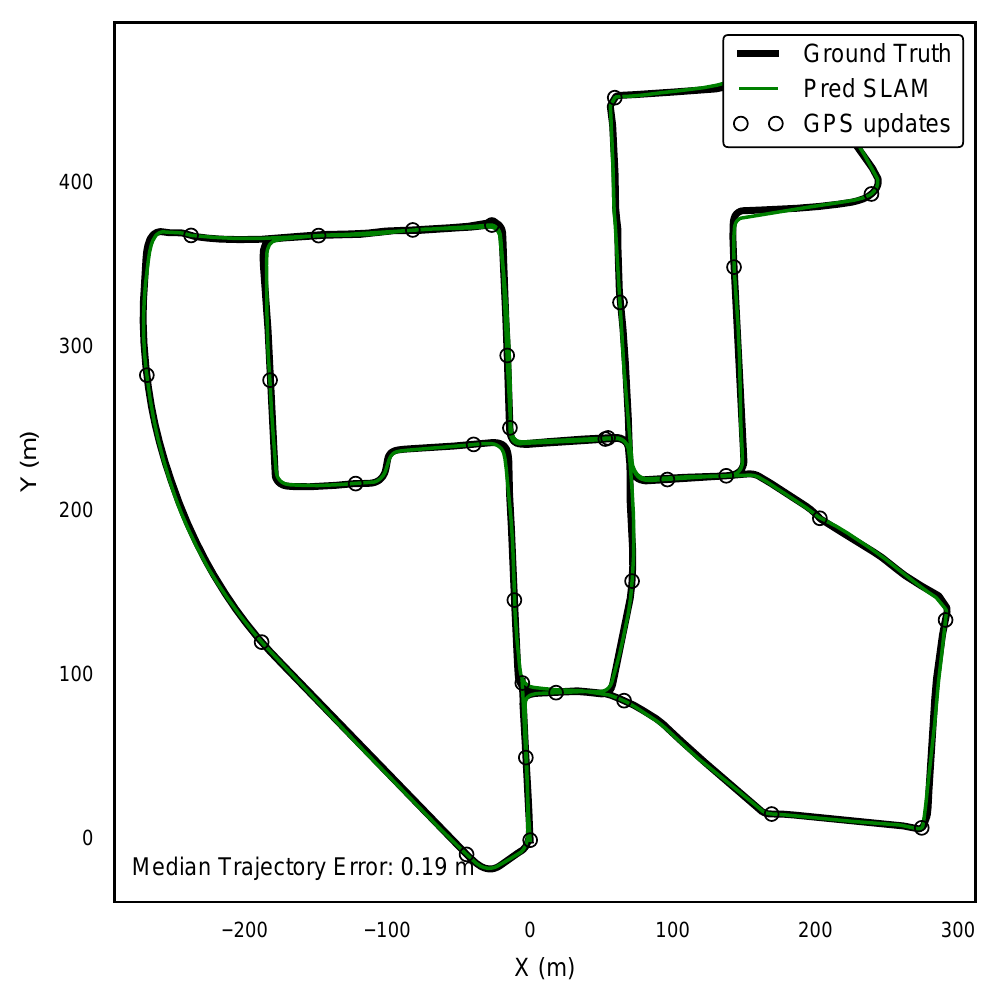}\\
        {\scriptsize \textbf{(a) Multi-FOV Synthetic Dataset}} &
                                                                 {\scriptsize
                                                                 \textbf{(b)
                                                                 Omnicam
                                                                 Dataset}}
        & {\scriptsize \textbf{(c) Oxford 1000km}}
        & {\scriptsize \textbf{(d) KITTI 00}}\vspace{1mm}\\
        \includegraphics[height=1.55in]{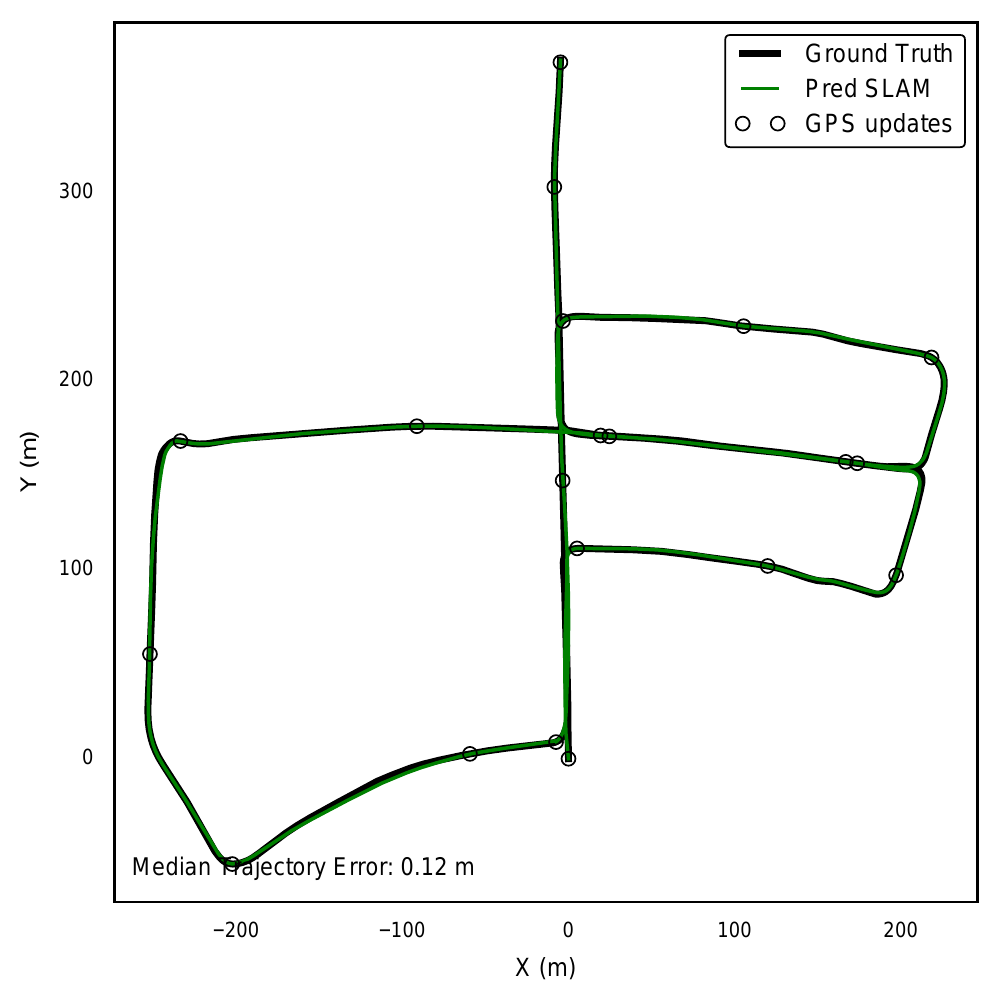}
        &\includegraphics[height=1.55in]{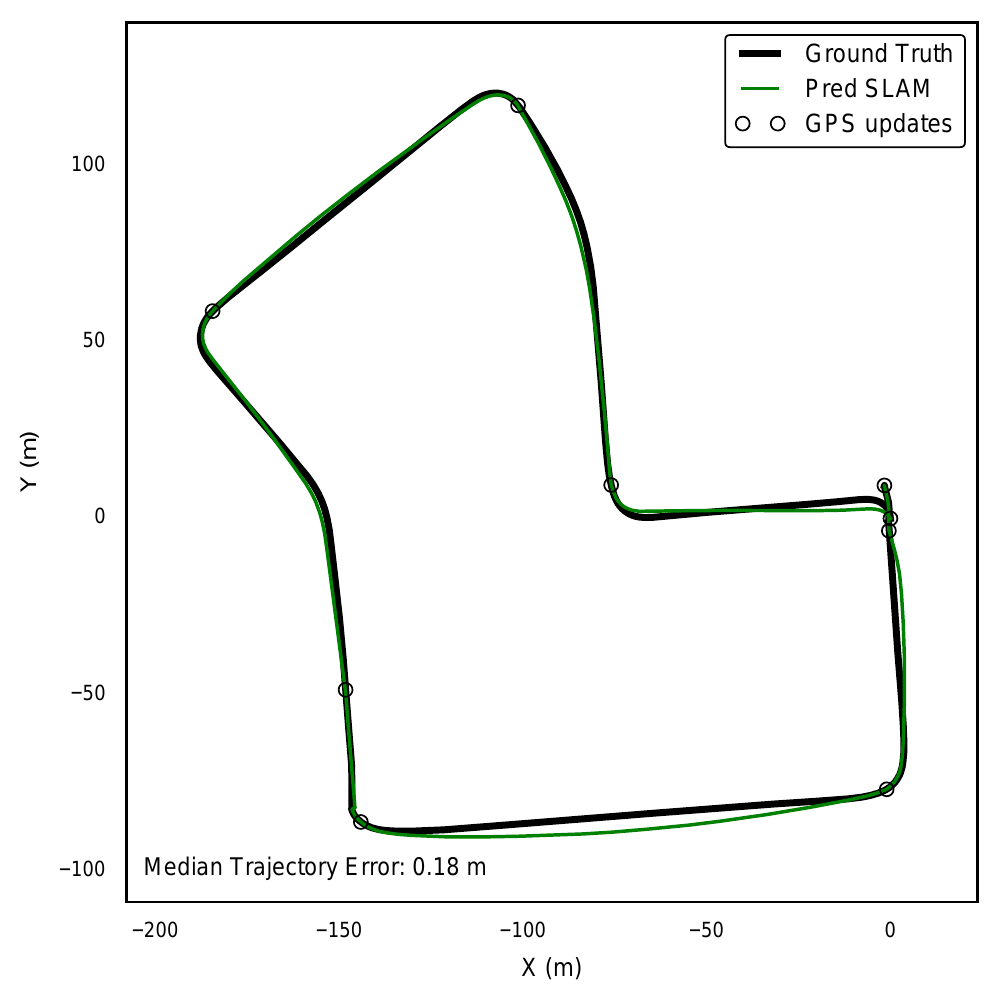}
        &\includegraphics[height=1.55in]{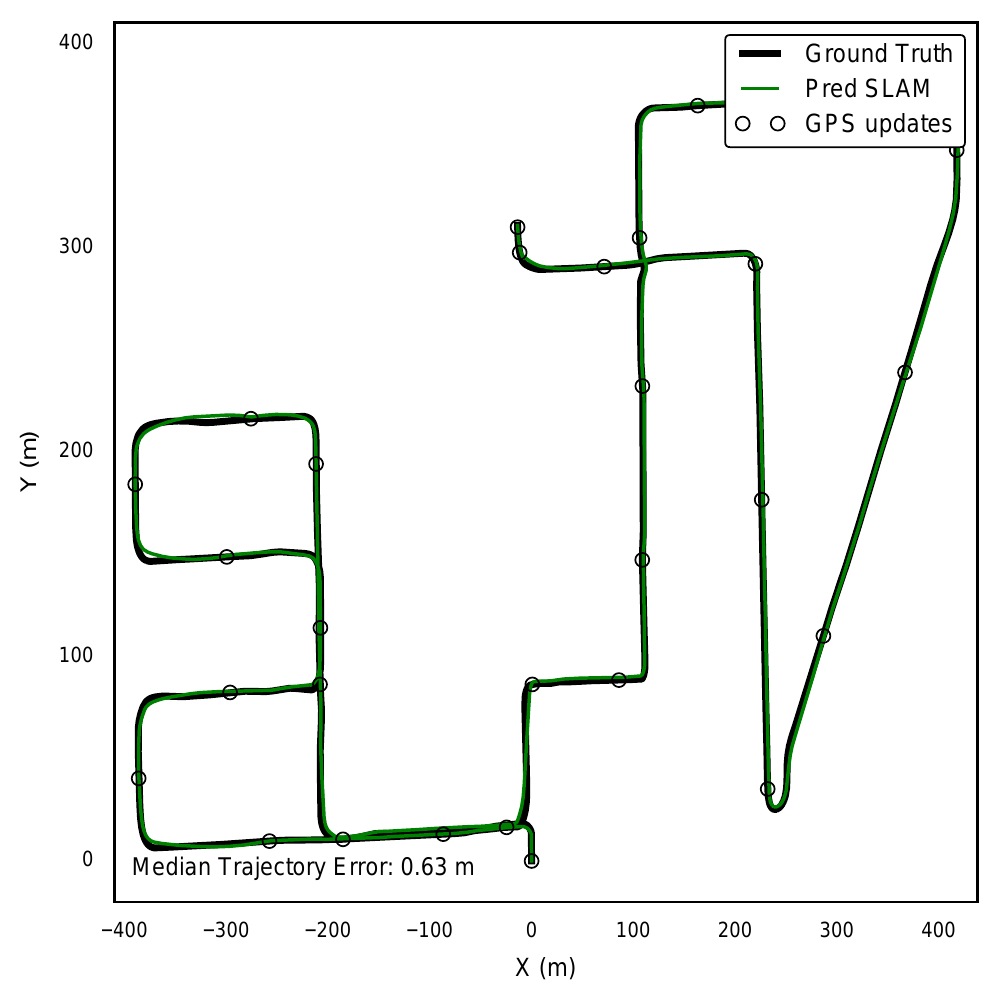}
        &\includegraphics[height=1.55in]{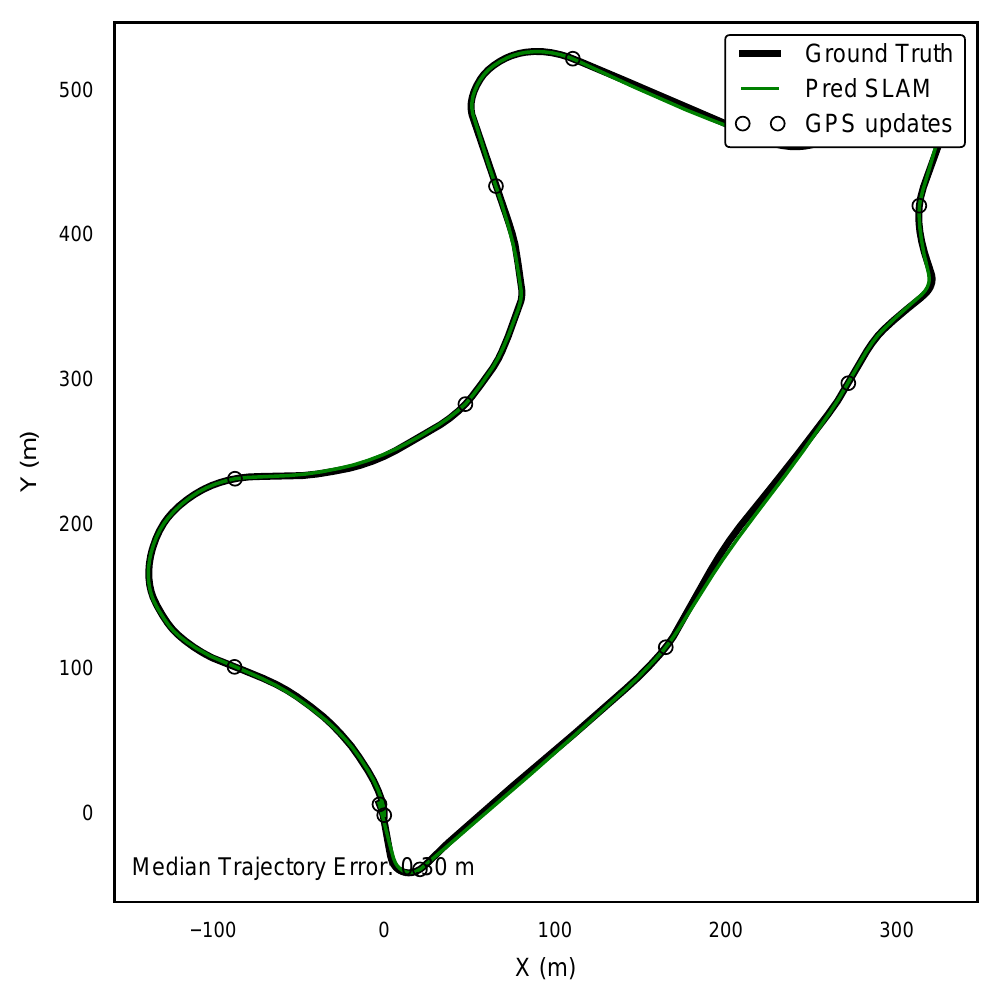}\\
        {\scriptsize \textbf{(e) KITTI 05}}  & {\scriptsize \textbf{(f) KITTI 07}} & {\scriptsize \textbf{(g) KITTI 08}} & {\scriptsize \textbf{(h) KITTI 09}}
      \end{tabular}}}
  \caption{\textbf{Sensor fusion with learned ego-motion}: On fusing
    our proposed VO method with intermittent GPS updates (every 150
    frames, black circles), the pose-graph optimized ego-motion solution
    (in green) achieves sufficiently high accuracy relative to ground
    truth. We test on a variety of publicly-available datasets including
    \textbf{(a)} Multi-FOV synthetic dataset~\cite{Zhang2016ICRA}
    (\textit{pinhole} shown above), \textbf{(b)} an omnidirectional-camera
    dataset~\cite{schonbein2014omnidirectional}, \textbf{(c)} Oxford Robotcar 1000km
    Dataset~\cite{maddern20161} (2015-11-13-10-28-08) \textbf{(d-h)} KITTI
    dataset~\cite{Geiger2012CVPR}. \textit{Weak supervision} such as GPS
    measurements can be especially advantageous in recovering improved
    estimates for localization, while simultaneously minimizing
    uncertainties associated with pure VO-based approaches. }
  \label{fig:egomotion-fusion}
  \vspace{-2mm}
\end{figure*}


\subsection{Evaluating ego-motion performance with sensor fusion}
\label{sec:performance} In this section, we evaluate our approach
against a few state-of-the-art algorithms for monocular visual
odometry~\cite{kitt2010visual}. On the KITTI
dataset~\cite{Geiger2012CVPR}, the pre-trained estimator is used to
robustly and accurately predict ego-motion from KLT features tracked
over the dataset image sequence. The frame-to-frame ego-motion
estimates are integrated for each session to recover the full
trajectory estimate and simultaneously fused with intermittent GPS
updates (incorporated every 150 frames). In
Figure~\ref{fig:egomotion-fusion}, we show the qualitative performance
in the overall trajectory obtained with our method.  The entire
pose-optimized trajectory is compared against the ground truth
trajectory. The
translational errors are computed for each of the ground truth and
prediction pose pairs, and their median value is reported in
Table~\ref{tab:trajectory-errors} for a variety of datasets with
varied camera optics.

%

\subsection{Varied camera optics}\label{sec:model-free} 
\begin{figure}[h]
  \centering 
  {\renewcommand{\arraystretch}{0.1} 
    {\setlength{\tabcolsep}{0.1mm}
      \begin{tabular}{ccc}
        \includegraphics[width=0.33\columnwidth]{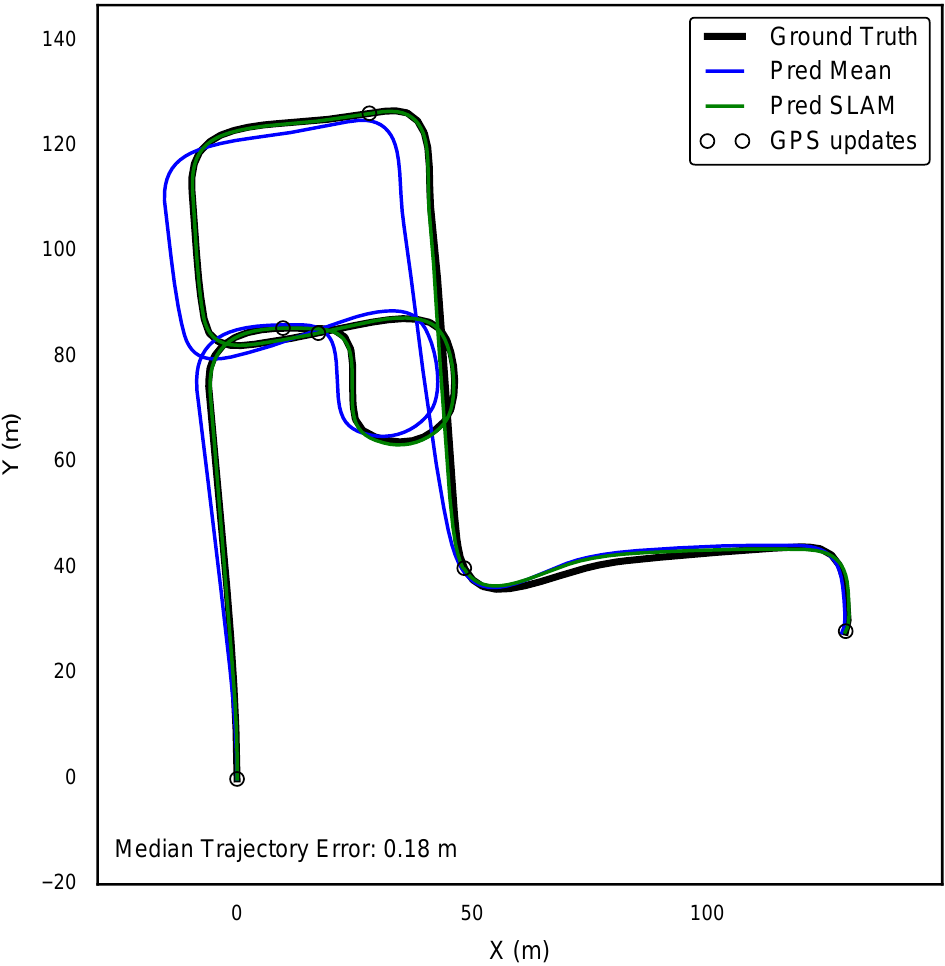}&
        \includegraphics[width=0.33\columnwidth]{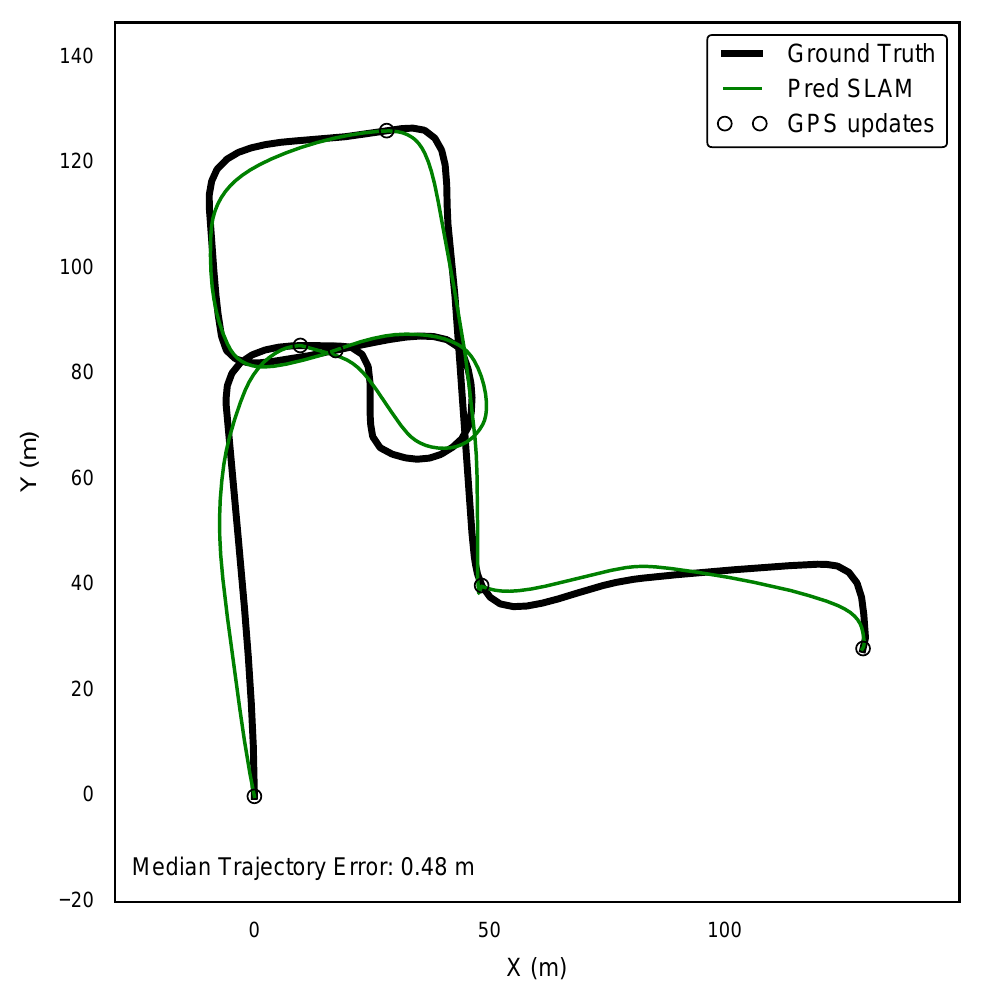}&
        \includegraphics[width=0.33\columnwidth]{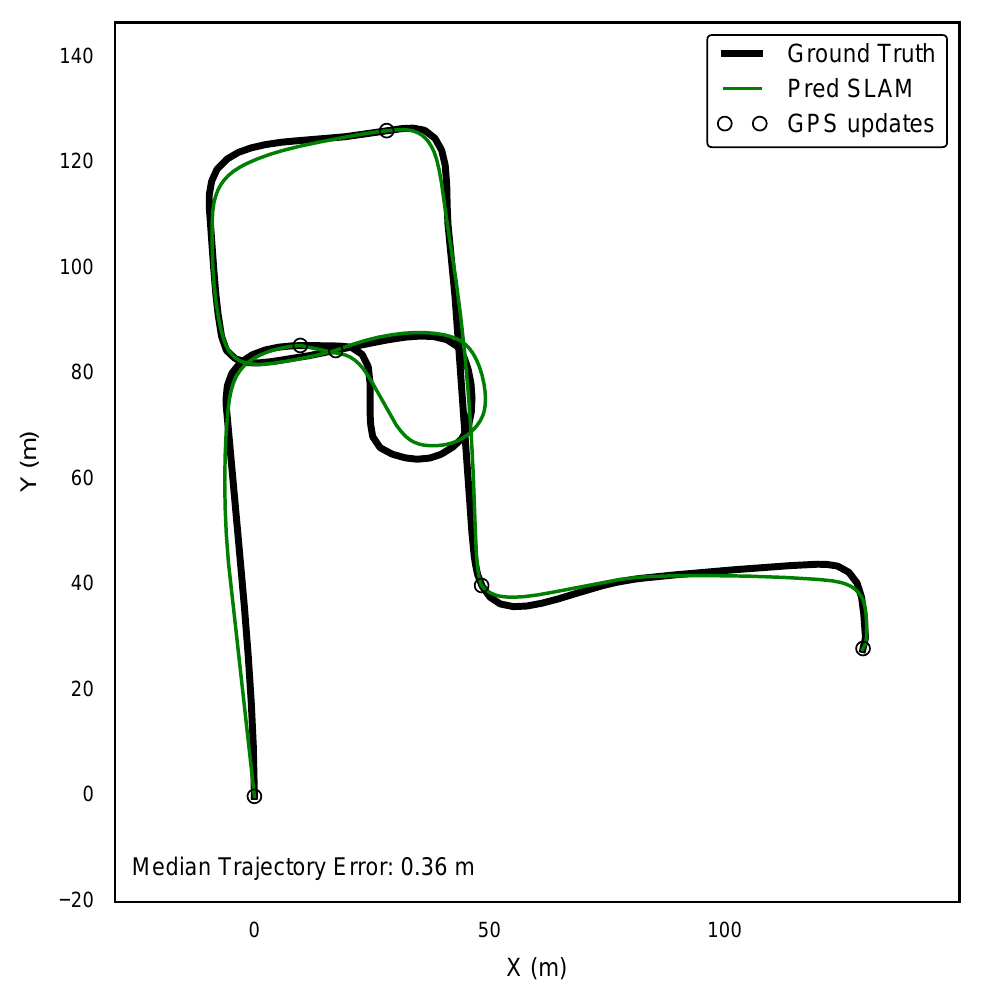}\\
        \scriptsize \textbf{Pinhole} & \scriptsize \textbf{Fisheye} & \scriptsize \textbf{Catadioptric}
      \end{tabular}}}
  \caption{\textbf{Varied camera optics:} An illustration of the
    performance of our general-purpose approach for varied camera optics
    (pinhole, fisheye, and catadioptric lenses) on the
    Multi-FOV synthetic dataset~\cite{Zhang2016ICRA}. Without any prior
    knowledge on the camera optics, or the mounting configuration
    (extrinsics), we are able to robustly and accurately recover the full
    trajectory of the vehicle (with intermittent GPS updates every 500
    frames).}
  \label{fig:egomotion-evaluation-varied-optics}
  \vspace{-6mm}
\end{figure}

Most of the
existing implementations of VO estimation are restricted to a class of
camera optics, and generally avoid implementing a general-purpose VO estimator for
varied camera optics. Our approach on the other hand, has shown the ability to provide
accurate VO with intermittent GPS trajectory estimation while
simultaneously being applicable to a varied range of camera
models. In
Figure~\ref{fig:egomotion-evaluation-varied-optics}, we compare with 
intermittent GPS trajectory estimates for all three camera models, and
verify their performance accuracy compared to ground truth. In our
experiments, we found that while our proposed solution was
sufficiently powerful to model different camera optics, it was
significantly better at modeling pinhole lenses as compared to fisheye
and catadioptric cameras (See Table~\ref{tab:trajectory-errors}). In
future work, we would like to investigate further extensions that
improve the accuracy for both fisheye and catadioptric lenses.

%

\begin{table}[!t]
\centering
\scriptsize
\rowcolors{2}{gray!25}{white}
\begin{tabular}{lM{1cm}M{2cm}}
\toprule
  \textbf{Dataset} & \textbf{Camera Optics} & \textbf{Median Trajectory Error} \\
  \midrule
  KITTI-00 & Pinhole & 0.19 m \\
  KITTI-02 & Pinhole & 0.30 m \\
  KITTI-05 & Pinhole & 0.12 m \\
  KITTI-07 & Pinhole & 0.18 m \\
  KITTI-08 & Pinhole & 0.63 m \\
  KITTI-09 & Pinhole & 0.30 m \\
  Multi-FOV~\cite{Zhang2016ICRA} & Pinhole & 0.18 m \\
  Multi-FOV~\cite{Zhang2016ICRA} & Fisheye & 0.48 m \\
  Multi-FOV~\cite{Zhang2016ICRA} & Catadioptric & 0.36 m \\
  Omnidirectional~\cite{schonbein2014omnidirectional} & Catadioptric & 0.52 m \\
  Oxford 1000km$^\dagger$~\cite{maddern20161} & Pinhole & 0.03 m \\
\bottomrule %
\end{tabular}\\\vspace{1mm}
\caption{\textbf{Trajectory prediction performance}: An illustration
  of the trajectory prediction performance of our proposed ego-motion
  approach when fused with intermittent GPS updates (every 150
  frames). The errors are computed
  across the entire length of the optimized trajectory and ground
  truth. For Oxford 1000km dataset, we only evaluate on a single
  session~\scriptsize{(2015-11-13-10-28-08 [80GB]: $^\dagger$Stereo Centre)}}.  
\label{tab:trajectory-errors}
\vspace{-9mm}
\end{table}

%

%
\subsection{Self-supervised Visual Ego-motion Learning in Robots}
\label{sec:bootstrap-exp} We
envision the capability of robots to self-supervise tasks such as
visual ego-motion estimation to be especially beneficial in the
context of life-long learning and autonomy. We experiment and validate
this concept through a concrete example using the 1000km Oxford Robot
Car dataset~\cite{maddern20161}. We train the task of visual
ego-motion on a new camera sensor by leveraging the fused GPS and INS
information collected on the robot car as ground truth trajectories
(6-DOF), and extracting feature trajectories (via KLT) from image
sequences obtained from the new
camera sensor. The timestamps from the cameras are synchronized
with respect to the timestamps of the fused GPS and INS information,
in order to obtain a one-to-one mapping for training purposes. We
train on the \texttt{stereo\_centre} \textit{(pinhole)} camera dataset
and present our results in Table~\ref{tab:trajectory-errors}. As seen
in Figure~\ref{fig:egomotion-fusion}, we are able to achieve
considerably accurate long-term state estimates by fusing our proposed
visual ego-motion estimates with even sparser GPS updates (every 2-3
seconds, instead of 50Hz GPS/INS readings). This allows the robot to
reduce its reliance on GPS/INS alone to perform robust, long-term
trajectory estimation.


%


%

\subsection{Implementation Details}\label{sec:implementation}
In this section we describe the details of our proposed model,
training methodology and parameters used. The input $\Bx = (\vx,
\Delta\vx)$ to the density-based ego-motion estimator are feature
tracks extracted via (Kanade-Lucas-Tomasi) KLT feature tracking over
the raw camera image sequences. The input feature positions and flow
vectors are normalized to be the in range of $[-1,1]$ using the
dimensions of the input image. We evaluate sparse LK (Lucas-Kanade)
optical flow over 7 pyramidal scales with a scale factor of
$\sqrt{2}$. As the features are extracted, the corresponding robot
pose (either available via GPS or GPS/INS/wheel odometry sensor
fusion) is synchronized and recorded in $SE(3)$ for training
purposes. The input KLT features, and the corresponding relative pose
estimates used for training are parameterized as $\mathbf{\Bz} =
(\mathbf{t},\mathbf{r}) \in \mathbb{R}^6$, with a Euclidean
translation vector $\mathbf{t} \in \mathbb{R}^3$ and an Euler rotation
vector $\mathbf{r} \in \mathbb{R}^3$.

\textbf{Network and training:} The proposed architecture
consists of a set of fully-connected stacked layers (with 1024, 128
and 32 units) followed by a Mixture Density Network with 32 hidden
units and 5 mixture components ($K$). Each of the initial
fully-connected layers implement \textit{tanh} activation after it,
followed by a dropout layer with a dropout rate of 0.1. The final
output layer of the MDN ($a^{\pi}$, $a^{\mu}$,
$a^{\sigma}$) consists of $(O + 2) * K$ outputs where $O$ is the
desired number of states estimated. 

The network is trained (in Stage 1) with loss weights of 10, 0.1, 1
corresponding to the losses $\mathcal{L}_{MDN}, \mathcal{L}_{TRAJ},
\mathcal{L}_{CVAE}$ described in previous sections. The training data
is provided in batches of 100 frame-to-frame subsequent image pairs,
each consisting of approximately 50 randomly sampled feature matches
via KLT. The learning rate is set to $1\mathrm{e}{-3}$ with Adam as
the optimizer. On the synthetic Multi-FOV dataset and the KITTI
dataset, training for most models took roughly an hour and a half
(3000 epochs) independent of the KLT feature extraction step.

\textbf{Two-stage optimization}: We found the one-shot joint
optimization of the \textit{local} ego-motion estimation and
\textit{global} trajectory optimization to have sufficiently low
convergence rates during training. One possible explanation is the
high sensitivity of the loss weight parameters that is used for tuning
the local and global losses into a single objective. As previously addressed in
Section~\ref{sec:trajectory-optimization}, we
separate the training into two stages thereby alleviating the
aforementioned issues, and maintaining fast convergence rates in Stage
1. Furthermore, we note that during the second stage, it only requires
a few tens of iterations for sufficiently accurate ego-motion
trajectories.  In order to optimize over a larger time-window in stage
2, we set the batch size to 1000 frame-to-frame image matches, again
randomly sampled from the training set as before. Due to the large integration
window and memory limitations, we train this stage purely on the CPU
for only 100 epochs each taking roughly 30s per epoch. Additionally,
in stage 2, the loss weights for $\mathcal{L}_{TRAJ}$ are increased to
100 in order to have faster convergence to the \textit{global}
trajectory. The remaining loss weights are left unchanged.

\textbf{Trajectory fusion}: We use
GTSAM\footnote{\scriptsize\url{http://collab.cc.gatech.edu/borg/gtsam}}
to construct the underlying factor graph for pose-graph
optimization. Odometry constraints obtained from the frame-to-frame
ego-motion are incorporated as a 6-DOF constraint parameterized in $SE(3)$ with $1*10^{-3}$~rad rotational noise
and $5*10^{-2}$~m translation noise. As with typical autonomous
navigation solutions, we expect measurement updates in the form of GPS
(absolute reference updates) in order to correct for the long-term
drift incurred in open-loop odometry chains. We incorporate absolute
prior updates only every 150 frames, with a weak translation prior of $0.01$~m. The constraints
are incrementally added and solved using iSAM2~\cite{kaess2012isam2} as
the measurements are streamed in, with updates performed every 10
frames.

While the proposed MDN is parametrized in Euler angles, the
\textit{trajectory integration module} parameterizes the rotation
vectors in quaternions for robust and unambiguous long-term trajectory
estimation. All the rigid body transformations are implemented
directly in Tensorflow for pure-GPU training support.

\textbf{Run-time performance}: We are particularly interested in the
run-time / test-time performance of our approach on CPU architectures
for mostly resource-constrained settings. Independent of the KLT feature
tracking run-time, we are able to recover ego-motion estimates at
roughly 3ms on a consumer-grade Intel(R) Core(TM)
i7-3920XM CPU @ 2.90GHz.


\textbf{Source code and Pre-trained weights}: We implemented the
MDN-based ego-motion estimator with Keras and Tensorflow, and trained
our models using a combination of CPUs and GPUs (NVIDIA Titan X). All
the code was trained on an server-grade Intel(R) Xeon(R) CPU E5-2630
v3 @ 2.40GHz and tested on the same consumer-grade machine as
mentioned above to
emulate potential real-world use-cases.  The source code and
pre-trained models used will be made available
shortly\footnote{\scriptsize
See~\url{http://people.csail.mit.edu/spillai/learning-egomotion}
and~\url{https://github.com/spillai/learning-egomotion}}. 


\section{Discussion}\label{sec:discussion} The initial results in
bootstrapped learning for visual ego-motion has motivated new
directions towards life-long learning in autonomous robots. While our
visual ego-motion model architecture is shown to be sufficiently
powerful to recover ego-motions for non-linear camera optics such as
fisheye and catadioptric lenses, we continue to investigate further
improvements to match existing state-of-the-art models for these lens
types. Our current model does not capture distortion effects yet,
however, this is very much a future direction we would like to
take. Another consideration is the resource-constrained setting, where
the optimization objective incorporates an additional regularization
term on the number of parameters used, and the computation load
consumed. We hope for this resource-aware capability to transfer to
real-world limited-resource robots and to have a significant impact on
the adaptability of robots for long-term autonomy.

\section{Conclusion}\label{sec:conclusion} While many visual
ego-motion algorithm variants have been proposed in the past decade,
we envision that a fully end-to-end trainable algorithm for generic
camera ego-motion estimation shall have far-reaching implications in
several domains, especially autonomous systems. Furthermore, we expect
our method to seamlessly operate under resource-constrained situations
in the near future by leveraging existing solutions in model reduction
and dynamic model architecture tuning. With the availability of
multiple sensors on these autonomous systems, we also foresee our
approach to bootstrapped task (visual ego-motion) learning to
potentially enable robots to learn from experience, and use the new
models learned from these experiences to encode redundancy and
fault-tolerance all within the same framework.


{
  \footnotesize
  \bibliographystyle{unsrt}
  \bibliography{tex/references}

\begin{thebibliography}{10}

\bibitem{nister2004visual}
David Nist{\'e}r, Oleg Naroditsky, and James Bergen.
\newblock Visual odometry.
\newblock In {\em Computer Vision and Pattern Recognition, 2004. CVPR 2004.
  Proceedings of the 2004 IEEE Computer Society Conference on}, volume~1, pages
  I--652. IEEE, 2004.

\bibitem{konolige2010large}
Kurt Konolige, Motilal Agrawal, and Joan Sola.
\newblock Large-scale visual odometry for rough terrain.
\newblock In {\em Robotics research}, pages 201--212. Springer, 2010.

\bibitem{howard2008real}
Andrew Howard.
\newblock Real-time stereo visual odometry for autonomous ground vehicles.
\newblock In {\em 2008 IEEE/RSJ International Conference on Intelligent Robots
  and Systems}, pages 3946--3952. IEEE, 2008.

\bibitem{kitt2010visual}
Bernd Kitt, Andreas Geiger, and Henning Lategahn.
\newblock Visual odometry based on stereo image sequences with {RANSAC}-based
  outlier rejection scheme.
\newblock In {\em Intelligent Vehicles Symposium}, pages 486--492, 2010.

\bibitem{hee2013motion}
Gim Hee~Lee, Friedrich Faundorfer, and Marc Pollefeys.
\newblock Motion estimation for self-driving cars with a generalized camera.
\newblock In {\em Proceedings of the IEEE Conference on Computer Vision and
  Pattern Recognition}, pages 2746--2753, 2013.

\bibitem{kneip2013using}
Laurent Kneip, Paul Furgale, and Roland Siegwart.
\newblock Using multi-camera systems in robotics: Efficient solutions to the
  {n-PnP} problem.
\newblock In {\em Robotics and Automation (ICRA), 2013 IEEE International
  Conference on}, pages 3770--3776. IEEE, 2013.

\bibitem{scaramuzza20111}
Davide Scaramuzza.
\newblock 1-point-{RANSAC} structure from motion for vehicle-mounted cameras by
  exploiting non-holonomic constraints.
\newblock {\em Int'l J.\ of Computer Vision}, 95(1):74--85, 2011.

\bibitem{triggs1999bundle}
Bill Triggs, Philip~F McLauchlan, Richard~I Hartley, and Andrew~W Fitzgibbon.
\newblock Bundle adjustment — {A modern synthesis}.
\newblock In {\em International workshop on vision algorithms}, pages 298--372.
  Springer, 1999.

\bibitem{hartley2003multiple}
Richard Hartley and Andrew Zisserman.
\newblock {\em Multiple view geometry in computer vision}.
\newblock Cambridge university press, 2003.

\bibitem{moravec1980obstacle}
Hans~P Moravec.
\newblock Obstacle avoidance and navigation in the real world by a seeing robot
  rover.
\newblock Technical report, DTIC Document, 1980.

\bibitem{matthies1989dynamic}
Larry~Henry Matthies.
\newblock Dynamic stereo vision.
\newblock 1989.

\bibitem{olson2000robust}
Clark~F Olson, Larry~H Matthies, H~Schoppers, and Mark~W Maimone.
\newblock Robust stereo ego-motion for long distance navigation.
\newblock In {\em Computer Vision and Pattern Recognition, 2000. Proceedings.
  IEEE Conference on}, volume~2, pages 453--458. IEEE, 2000.

\bibitem{fischler1981random}
Martin~A Fischler and Robert~C Bolles.
\newblock Random sample consensus: A paradigm for model fitting with
  applications to image analysis and automated cartography.
\newblock {\em Communications of the ACM}, 24(6):381--395, 1981.

\bibitem{corke2004omnidirectional}
Peter Corke, Dennis Strelow, and Sanjiv Singh.
\newblock Omnidirectional visual odometry for a planetary rover.
\newblock In {\em Intelligent Robots and Systems, 2004.(IROS 2004).
  Proceedings. 2004 IEEE/RSJ International Conference on}, volume~4, pages
  4007--4012. IEEE, 2004.

\bibitem{liang2002visual}
Bojian Liang and Nick Pears.
\newblock Visual navigation using planar homographies.
\newblock In {\em Robotics and Automation, 2002. Proceedings. ICRA'02. IEEE
  International Conference on}, volume~1, pages 205--210. IEEE, 2002.

\bibitem{ke2003transforming}
Qifa Ke and Takeo Kanade.
\newblock Transforming camera geometry to a virtual downward-looking camera:
  Robust ego-motion estimation and ground-layer detection.
\newblock In {\em Computer Vision and Pattern Recognition, 2003. Proceedings.
  2003 IEEE Computer Society Conference on}, volume~1, pages I--390. IEEE,
  2003.

\bibitem{scaramuzza2009real}
Davide Scaramuzza, Friedrich Fraundorfer, and Roland Siegwart.
\newblock Real-time monocular visual odometry for on-road vehicles with 1-point
  {RANSAC}.
\newblock In {\em Robotics and Automation, 2009. ICRA'09. IEEE International
  Conference on}, pages 4293--4299. IEEE, 2009.

\bibitem{roberts2009learning}
Richard Roberts, Christian Potthast, and Frank Dellaert.
\newblock Learning general optical flow subspaces for egomotion estimation and
  detection of motion anomalies.
\newblock In {\em Computer Vision and Pattern Recognition, 2009. CVPR 2009.
  IEEE Conference on}, pages 57--64. IEEE, 2009.

\bibitem{ciarfuglia2014evaluation}
Thomas~A Ciarfuglia, Gabriele Costante, Paolo Valigi, and Elisa Ricci.
\newblock Evaluation of non-geometric methods for visual odometry.
\newblock {\em Robotics and Autonomous Systems}, 62(12):1717--1730, 2014.

\bibitem{costante2016exploring}
Gabriele Costante, Michele Mancini, Paolo Valigi, and Thomas~A Ciarfuglia.
\newblock {Exploring Representation Learning With {CNN}s for Frame-to-Frame
  Ego-Motion Estimation}.
\newblock {\em IEEE Robotics and Automation Letters}, 1(1):18--25, 2016.

\bibitem{Geiger2012CVPR}
Andreas Geiger, Philip Lenz, and Raquel Urtasun.
\newblock Are we ready for autonomous driving? {The KITTI} vision benchmark
  suite.
\newblock In {\em Proc.\ IEEE Conf.\ on Computer Vision and Pattern Recognition
  (CVPR)}, 2012.

\bibitem{wen2016vinet}
Ronald Clark, Sen Wang, Hongkai Wen, Andrew Markham, and Niki Trigoni.
\newblock {VINet}: {V}isual-{I}nertial odometry as a sequence-to-sequence
  learning problem.
\newblock AAAI, 2016.

\bibitem{scaramuzza2011visual}
Davide Scaramuzza and Friedrich Fraundorfer.
\newblock Visual odometry [tutorial].
\newblock {\em IEEE robotics \& automation magazine}, 18(4):80--92, 2011.

\bibitem{konda2015learning}
Kishore Konda and Roland Memisevic.
\newblock Learning visual odometry with a convolutional network.
\newblock In {\em International Conference on Computer Vision Theory and
  Applications}, 2015.

\bibitem{birchfield2007klt}
Stan Birchfield.
\newblock {KLT}: An implementation of the {Kanade-Lucas-Tomasi} feature
  tracker, 2007.

\bibitem{bishop1994mixture}
Christopher~M Bishop.
\newblock {Mixture Density Networks}.
\newblock 1994.

\bibitem{Zhang2016ICRA}
Zichao Zhang, Henri Rebecq, Christian Forster, and Davide Scaramuzza.
\newblock Benefit of large field-of-view cameras for visual odometry.
\newblock In {\em IEEE International Conference on Robotics and Automation
  (ICRA)}. IEEE, 2016.

\bibitem{schonbein2014omnidirectional}
Miriam Sch{\"o}nbein and Andreas Geiger.
\newblock Omnidirectional 3d reconstruction in augmented manhattan worlds.
\newblock In {\em Intelligent Robots and Systems (IROS 2014), 2014 IEEE/RSJ
  International Conference on}, pages 716--723. IEEE, 2014.

\bibitem{maddern20161}
Will Maddern, Geoffrey Pascoe, Chris Linegar, and Paul Newman.
\newblock 1 year, 1000 km: {The Oxford RobotCar dataset}.
\newblock {\em Int'l J.\ of Robotics Research}, page 0278364916679498, 2016.

\bibitem{kaess2012isam2}
Michael Kaess, Hordur Johannsson, Richard Roberts, Viorela Ila, John~J Leonard,
  and Frank Dellaert.
\newblock {iSAM2}: Incremental smoothing and mapping using the bayes tree.
\newblock {\em Int'l J.\ of Robotics Research}, 31(2):216--235, 2012.

\end{thebibliography}
}

\end{document}